\documentclass[sigconf]{acmart}
\usepackage[font=small]{caption}
\usepackage[font=footnotesize]{subcaption}
\usepackage{xcolor}

\AtBeginDocument{%
  \providecommand\BibTeX{{%
    \normalfont B\kern-0.5em{\scshape i\kern-0.25em b}\kern-0.8em\TeX}}}

\setcopyright{acmlicensed} 

\copyrightyear{2019}
\acmYear{2019}
\acmConference[SUMAC '19]{1st Workshop on Structuring and Understanding of Multimedia heritAge Contents}{October 21, 2019}{Nice, France}
\acmBooktitle{1st Workshop on Structuring and Understanding of Multimedia heritAge Contents (SUMAC '19), October 21, 2019, Nice, France}
\acmPrice{15.00}
\acmDOI{10.1145/3347317.3357239}
\acmISBN{978-1-4503-6910-7/19/10}

\begin{document}
\fancyhead{} 
\title{Historical and Modern Features for Buddha Statue Classification}

\author{Benjamin Renoust, Matheus Oliveira Franca, Jacob Chan, Noa Garcia, Van Le, Ayaka Uesaka, Yuta Nakashima, Hajime Nagahara}
\email{renoust@ids.osaka-u.ac.jp}
\affiliation{%
  \institution{Institute for Datability Science, Osaka University}
  \city{Osaka}
  \country{Japan}
}

\author{Jueren Wang}
\author{Yutaka Fujioka}
\email{fujioka@let.osaka-u.ac.jp}
\affiliation{%
  \institution{Graduate School of Letters, Osaka University}
  \city{Osaka}
  \country{Japan}}

\renewcommand{\shortauthors}{Renoust, et al.}

\begin{abstract}
While Buddhism has spread along the Silk Roads, many pieces of art have been displaced. Only a few experts may identify these works, subjectively to their experience. The construction of Buddha statues was taught through the definition of canon rules, but the applications of those rules greatly varies across time and space. Automatic art analysis aims at supporting these challenges. We propose to automatically recover the proportions induced by the construction guidelines, in order to use them and compare between different deep learning features for several classification tasks, in a medium size but rich dataset of Buddha statues, collected with experts of Buddhism art history.

\end{abstract}

\ccsdesc[500]{Applied computing~Fine arts}
\ccsdesc[300]{Computing methodologies~Image representations}
\ccsdesc[300]{Computing methodologies~Interest point and salient region detections}

\keywords{Art History, Buddha statues, classification, face landmarks}

\begin{teaserfigure}
  \label{fig:teaser}
\end{teaserfigure}

\maketitle

\section{Introduction}

\begin{figure*}[htbp]
	\centering
	\includegraphics[width=\linewidth]{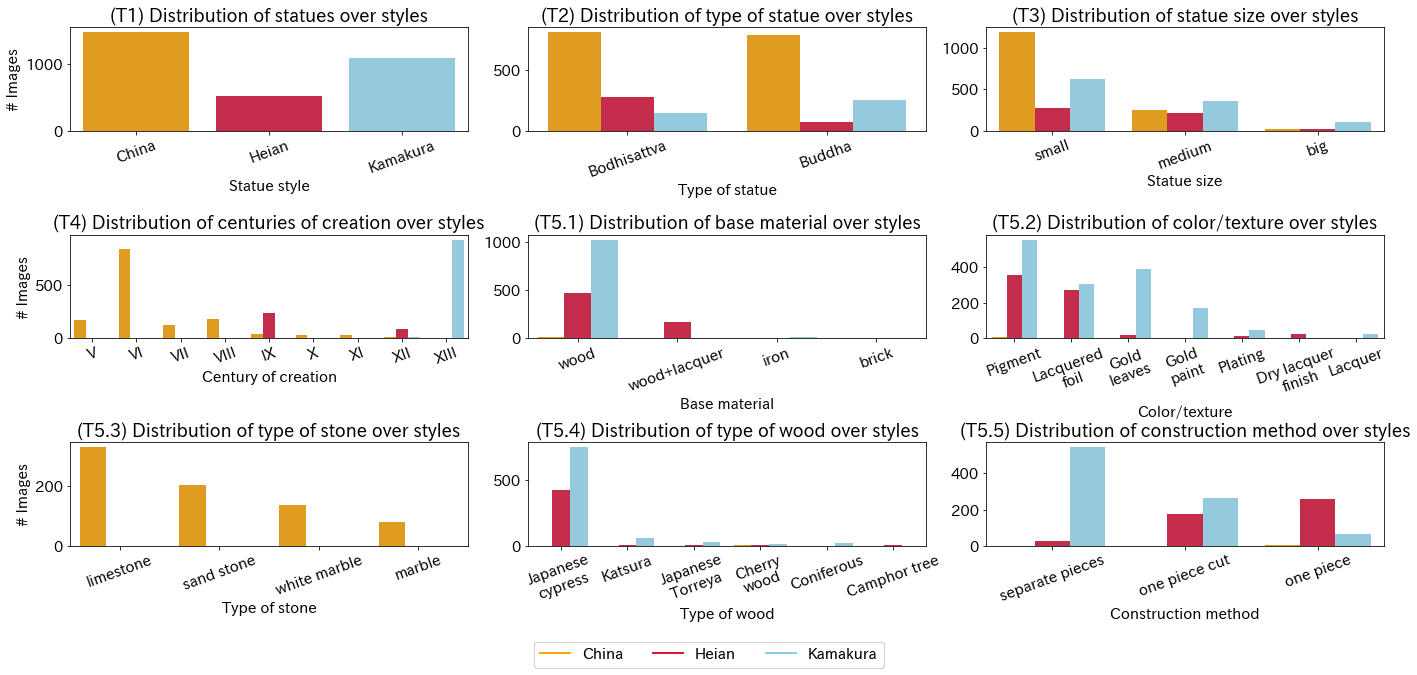}
	\caption{Distribution of all the images across the different classes with the period highlighted.}
	\label{fig:histoclasses}
\end{figure*}

Started in India, Buddhism spread across all the Asian subcontinent through China reaching the coasts of South-eastern Asia and the Japanese archipelago, benefiting from the travels along the Silk Roads~\cite{secret2006,faces2013}. The story is still subject to many debates as multiple theories are confronting on how this spread and evolution took place~\cite{spread1986, Guide, secret2006,faces2013}. Nonetheless, as Buddhism flourished along the centuries, scholars have exchanged original ideas that further diffused, shaping the different branches of Buddhism and art as we know them today. When Buddhism reached new territories, local people would craft Buddhism art themselves. Not only they observed common rules of crafting, but also they adapted them to express their own culture, giving rise to new styles~\cite{style1987}. 

With multiple crisis and cultural exchanges, many pieces of art have been displaced, and their sources may remain today uncertain. Only a few experts can identify these works. This is however subject to their own knowledge, and the origin of some statues is still disputed today~\cite{controversy2011}. However, our decade has seen tremendous progress in machine learning, so that we may harvest these techniques to support art identification~\cite{blessing2010using}.

\begin{figure*}[!ht]
	\centering
        \begin{subfigure}[b]{0.15\textwidth}
			\centering
			\includegraphics[height=3.8cm,]{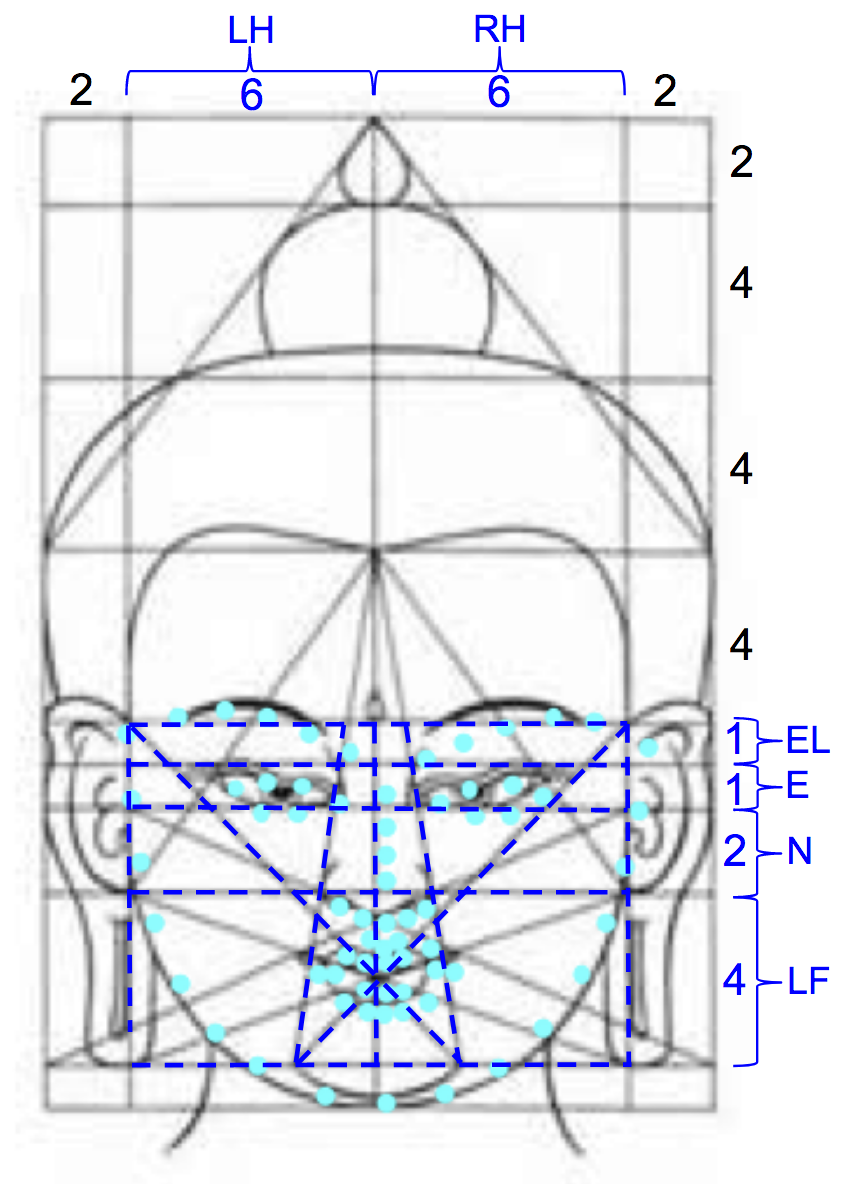}
			\caption{}
			\label{fig:tibetlines}
		\end{subfigure}
		\hspace*{-1.7mm}
		\begin{subfigure}[b]{0.24\textwidth}
			\centering
			\includegraphics[width=3.8cm,height=3.8cm,]{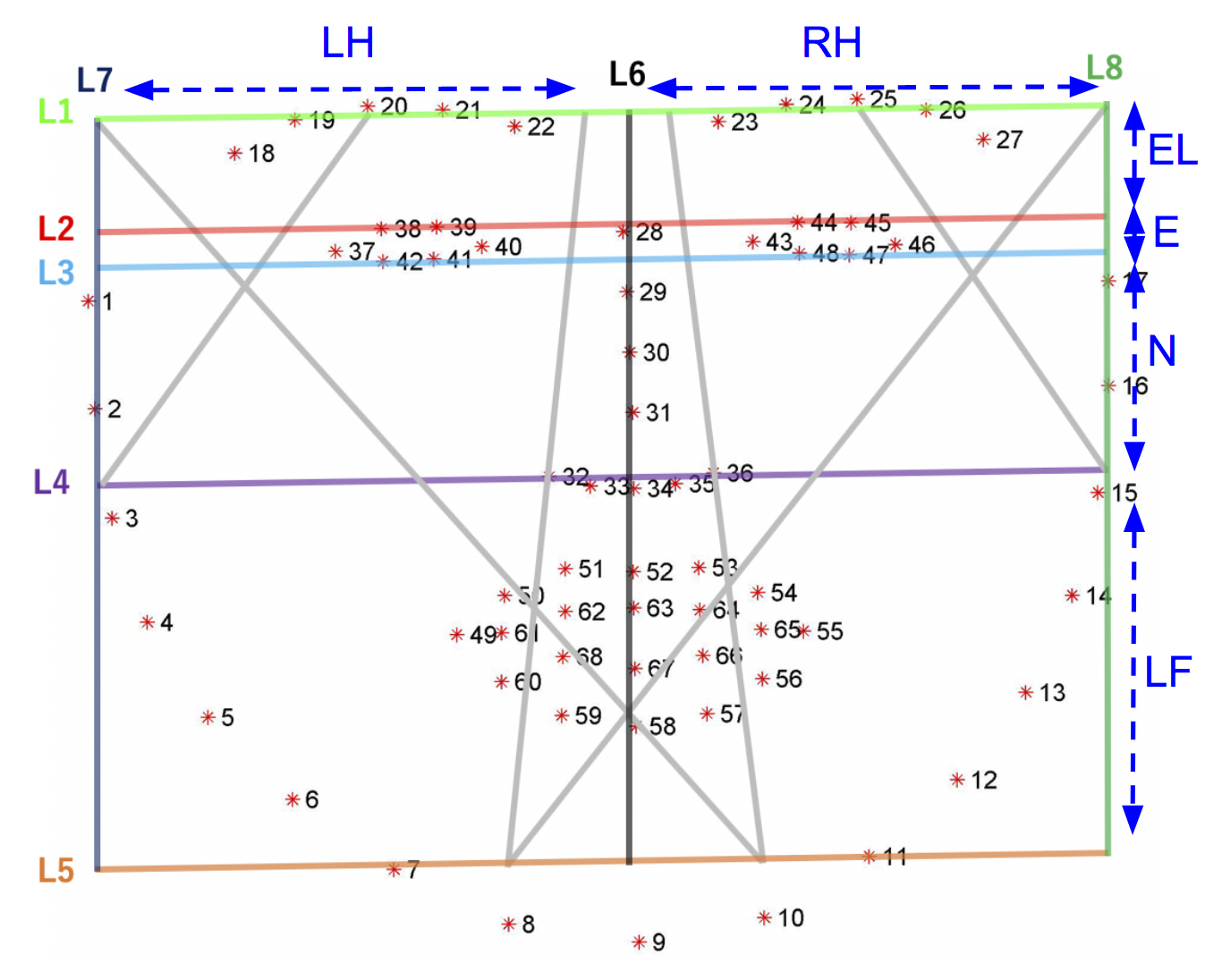}
			\caption{}
			\label{fig:linelabels}
		\end{subfigure}		
        \hspace*{-1.7mm}
		\begin{subfigure}[b]{0.22\textwidth}
			\centering
			\includegraphics[height=3.6cm]{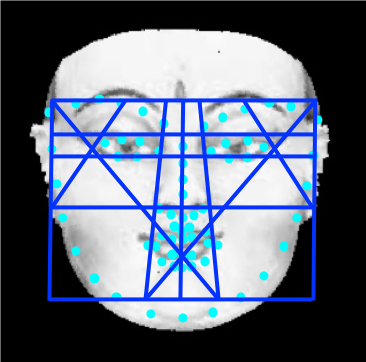}
			\caption{}
			\label{fig:model68landmarks}
		\end{subfigure}
		\hspace*{-1.8mm}
		\begin{subfigure}[b]{0.15\textwidth}
			\centering
			\includegraphics[height=3.6cm]{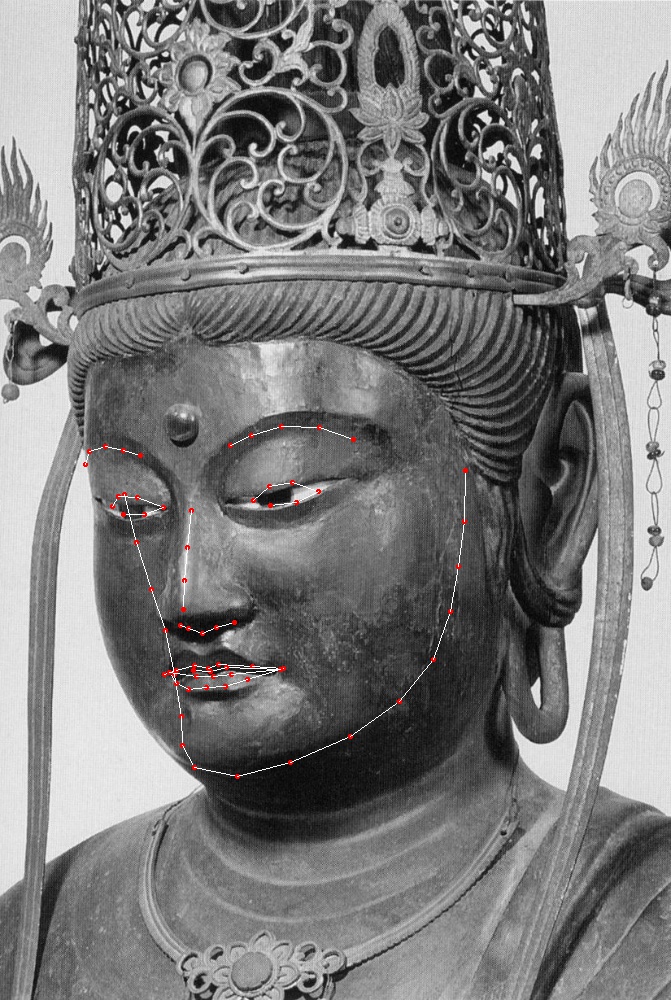}
			\caption{}
			\label{fig:buddha68landmarks}
		\end{subfigure}
		\hspace*{-1.8mm}
		\begin{subfigure}[b]{0.22\textwidth}
			\centering
			\includegraphics[height=3.6cm,]{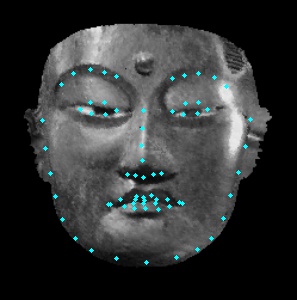}
			\caption{}
			\label{fig:buddha2D68frontal}
		\end{subfigure}
    \\ 
		\begin{subfigure}[b]{0.24\textwidth}
			\centering
			\includegraphics[width=4.2cm,height=4.0cm]{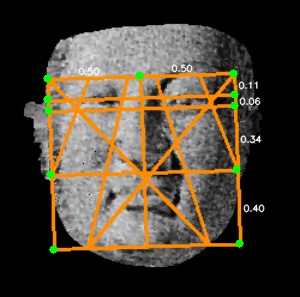}
			\caption{}
			\label{fig:lineschina}
		\end{subfigure}
	    \begin{subfigure}[b]{0.24\textwidth}
			\centering
			\includegraphics[width=4.2cm,height=4.0cm,]{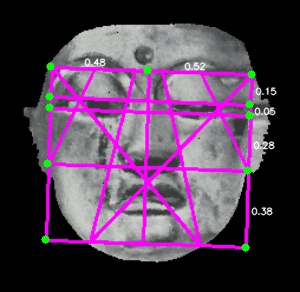}
			\caption{}
			\label{fig:linesheian}
		\end{subfigure}
		\begin{subfigure}[b]{0.24\textwidth}
			\centering
			\includegraphics[width=4.2cm,height=4.0cm]{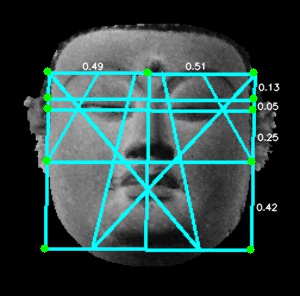}
			\caption{}
			\label{fig:lineskamakura}
		\end{subfigure}
		\begin{subfigure}[b]{0.24\textwidth}
			\centering
			\includegraphics[width=4.2cm,height=4.0cm]{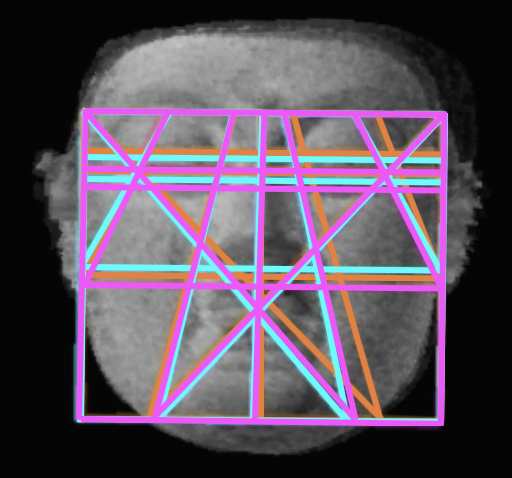}
			\caption{}
			\label{fig:linescombined}
		\end{subfigure} 

	\caption{Above: Deriving the Buddha iconometric proportions based on  68 facial landmarks.\ (a) Proportional measurements on a Tibetan canon of Buddha~\cite{tibetan} facial regions and their value (original template \textcopyright Carmen Mensik, \url{www.tibetanbuddhistart.com}).\ (b) Iconometric proportional guidelines defined from the 68 facial landmark points.\ (c) Application of landmarks and guidelines detection to the Tibetan model~\cite{tibetan}.\ (d) 3D 68 facial landmarks detected on a Buddha statue image, and its frontal projection (e). Below: Examples of the detected iconometric proportions in three different styles.\ (f) China.\ (g) Heian.\ (h) Kamakura.\ (i) The combined and superimposed iconometric proportional lines from the examples (f)-(h). Canon image is the courtesy of Carmen Mensink~\cite{tibetan}.\
	}
	\label{fig:buddhalandmarks}
\end{figure*}

Our work focuses on the representation of Buddha, central to Buddhism art, and more specifically on Buddha statues. Statues are 3D objects by nature. There are many types of Buddha statues, but all of them obey construction rules. These are \textit{canons}, that make a set of universal rules or principals to establish the very fundamentals of the representation of Buddha. Although the canons have first been taught using language-based description, these rules have been preserved today, and are consigned graphically in rule books (\textit{e.g.} Tibetan representations ~\cite{tibet17--, tibetan} as illustrated in Fig.~\ref{fig:tibetlines}). The study of art pieces measurements, or iconometry, may further be used to investigate the differences between classes of Buddha statues~\cite{reportya}.

In this paper, we are interested in understanding how these rules can reflect in a medium size set of Buddha statues (>1k identified statues in about 7k images). We focus in particular on the faces of the statues, through photographs taken from these statues (each being a 2D projection of the 3D statue). We propose to automatically recover the construction guidelines, and proceed with iconometry in a systematic manner. Taking advantage of the recent advances in image description, we further investigate different deep features and classification tasks of Buddha statues. This paper contributes by setting a baseline for the comparison between ``historical'' features, the set of canon rules, against ``modern'' features, on a medium size dataset of Buddha statues and pictures.

The rest of the paper is organized as follows. After discussing the related work, we present our dataset in Section~\ref{sec:data}. We then introduce the iconometry measurement and application in Section~\ref{sec:iconometry}. From this point on, we study different embedding techniques and compare them along a larger set of classification tasks (Sec.~\ref{sec:classification}) before concluding.



\subsection{Related Work}

Automatic art analysis is not a new topic, and early works have focused on hand crafted feature extraction to represent the content typically of paintings~\cite{johnson2008image,shamir2010impressionism,carneiro2012artistic,khan2014painting,mensink2014rijksmuseum}. These features were specific to their application, such as the author identification by brushwork decomposition using wavelets~\cite{johnson2008image}. A combination of color, edge, and texture features was used for author/school/style classification~\cite{shamir2010impressionism, khan2014painting}.
The larger task of painting classification has also been approached in a much more traditional way with SIFT features~\cite{carneiro2012artistic,mensink2014rijksmuseum}.

This was naturally extended to the use of deep visual features with great effectiveness~\cite{Bar2014ClassificationOA,karayev2014recognizing,Saleh2015LargescaleCO,elgammal2015quantifying, Tan2016CeciNP,ma2017part,mao2017deepart,elgammal2018shape,Garcia2018How,strezoski2018omniart}. The first approaches were using pre-trained networks for automatic classification~\cite{Bar2014ClassificationOA,karayev2014recognizing,Saleh2015LargescaleCO}. Fine tuned networks have then shown improved performances~\cite{Tan2016CeciNP,seguin2016visual,mao2017deepart,strezoski2017omniart, chu2018image}. Recent approaches~\cite{Garcia2018How, garcia2019context} introduced the combination of multimedia information in the form of joint visual and textual models~\cite{Garcia2018How} or using graph modeling~\cite{garcia2019context} for the semantic analysis of paintings. The analysis of style has also been investigated with relation to time and visual features~\cite{elgammal2015quantifying, elgammal2018shape}. Other alternatives are exploring domain transfer for object and face detection and recognition~\cite{crowley2015face,crowley2014state,crowley2016art}.

These methods mostly focus on capturing the visual content of paintings, on very well curated datasets. However, paintings are very different to Buddha statues, in that sense that statues are 3D objects, created with strict rules.
In addition, we are interested by studying the history of art, not limited to the visual appearance, but also about their historical, material, and artistic context. In this work, we explore different embeddings, from ancient Tibetan rules, to modern visual, in addition to face-based, and graph-based, for different classification tasks of Buddha statues.

We can also investigate recent works which are close to our application domain, \textit{i.e.} the analysis of oriental statues \cite{kamakura2005classification, ikeuchi2007great, reportya, bevan2014computer, bhaumik2018recognition, wang2019average}. Although, one previous work has achieved Thai statue recognition by using handcrafted facial features~\cite{pornpanomchai2011thai}. Other related works focus on the 3D acquisition of statues ~\cite{kamakura2005classification, ikeuchi2007great} and their structural analysis~\cite{bevan2014computer, bhaumik2018recognition}, with sometimes the goals of classification too~\cite{kamakura2005classification, reportya}.
We should also highlight the recent use of inpainting techniques on Buddhism faces for the study and recovery of damaged pieces~\cite{wang2019average}. 

Because 3D scanning does not scale to the order of thousands statues, we investigate features of 2D pictures of 3D statues, very close to the spirit of Pornpanomchai \textit{et al.}~\cite{pornpanomchai2011thai}. In addition to the study of ancient proportions, we provide modern analysis with visual, face-based (which also implies a 3D analysis), and semantic features for multiple classification tasks, on a very sparse dataset that does not provide information for every class.

\section{Data} \label{sec:data}

This work is led in collaboration with experts who wish to investigate three important styles of Buddha statues. A first style is made of statues from ancient \textbf{China} spreading between the IV and XIII centuries. A second style is made of Japanese statues during the \textbf{Heian} period (794-1185). The last style is also made of Japanese statues, during the \textbf{Kamakura} era (1185-1333). 

To do so, our experts have captured (scanned) photos in 4 series of books, resulting in a total of 6811 scanned images, and documented 1393 statues among them. The first series~\cite{chinabook} concerns 1076 Chinese statues (1524 pictures). Two book series~\cite{heian1,heian2} regroup 132 statues of the Heian period (1847 pictures). The last series~\cite{Kamakurabook} collects 185 statues of the Kamakura era (3888 pictures). 

To further investigate the statues, our experts have also manually curated extra meta-data information (only when available). For the \textbf{localization}, we so far only consider China and Japan. \textbf{Dimensions} are reporting the height of each statue, so we created three classes: \textit{small} (from 0 to 100 cm), \textit{medium} (from 100cm to 250cm) and \textit{big} (greater than 250 cm). Many statues also have a specific \textbf{statue type} attributed to them. We threshold them to the most common types, represented by at least 20 pictures, namely \textit{Bodhisattva} and \textit{Buddha}.

A \textbf{temporal information} which can be inferred from up to four components: an exact international date, a date or period that may be specific to the Japanese or Chinese traditional dating system, a century information (period), an era that may be specific to Japan or China (period). Because these information may be only periods, we re-align them temporally to give an estimate year in the international system, that is the median year of the intersection of all potential time periods. They all distribute between the V and XIII century.

\textbf{Material information} is also provided but it is made of multiple compounds and/or subdivisions. We observe the following categories: \textbf{base material} can be of \textit{wood}, \textit{wood+lacquer}, \textit{iron}, or \textit{brick}; \textbf{color or texture} can refer to \textit{pigment}, \textit{lacquered foil}, \textit{gold leaves}, \textit{gold paint}, \textit{plating}, \textit{dry lacquer finish}, or \textit{lacquer}; \textbf{type of stone} (when applies) may be \textit{limestone}, \textit{sand stone}, \textit{white marble}, or \textit{marble}; \textbf{type of wood} (also when applies) may be \textit{Japanese cypress},  \textit{Katsura}, \textit{Japanese Torreya}, \textit{cherry wood}, \textit{coniferous}, or \textit{camphor tree}; the material may also imply a \textbf{construction method} among \textit{separate pieces}, \textit{one piece cut}, and \textit{one piece}.

Fig.~\ref{fig:histoclasses} shows the distribution of all the images across the different classes. Because for each of the statues many information is either uncertain or unavailable, we can note that the data is very sparse, and most of the different classes are balanced unevenly. Note that not all pictures are corresponding to a documented statue, the curated dataset annotates a total of 3065 images in 1393 unique statues. In addition, not the same statues shares the same information, \textit{i.e.} some statues have color information, but no base material, when others have temporal information only \textit{etc.} As a consequence, each classification task we describe later in Sec.~\ref{sec:classification} has a specific subset of images and statues to which it may apply, not necessary overlapping with the subsets of other tasks.

\section{Iconometry}\label{sec:iconometry}

\begin{figure*}[ht]
	\centering
\includegraphics[width=\linewidth]{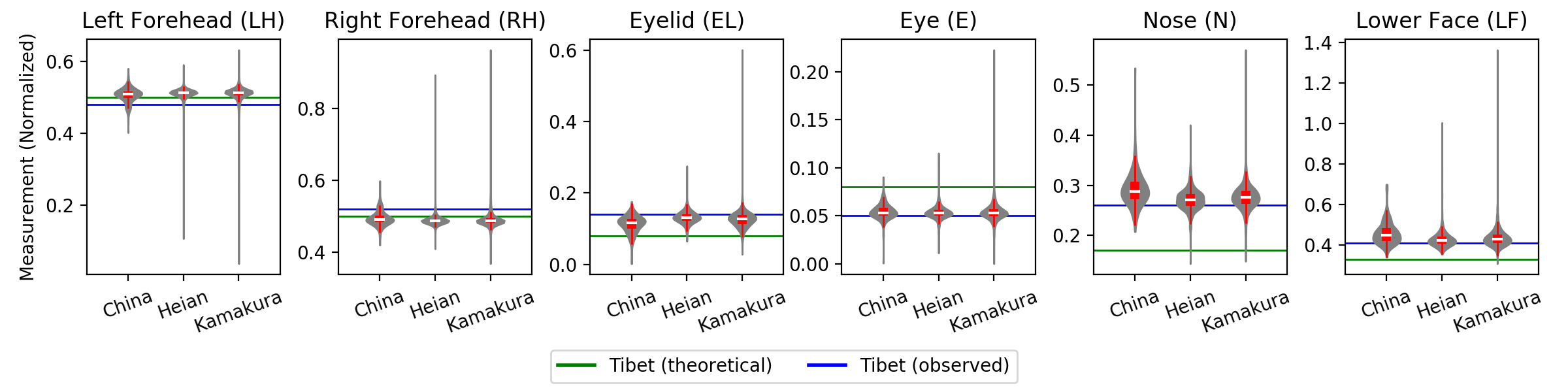}
	\caption{Six iconometric proportions distribution across the three styles, China, Kamakura, and Heian, against Tibetan theoretical canons and their actually observed proportions.}
	\label{fig:allhistorical}
\end{figure*}

We begin our analysis with the use of historic iconometry for determining facial proportions in figurative Buddha constructions. For this, we have chosen a model based on a Tibetan-originated 18th century book comprising of precise iconometric guidelines for representing Buddha-related artworks~\cite{tibet17--, tibetan}. Although this book primarily encompasses Tibetan-based Buddha drawing gui\-delines, it gave us insights of how Buddha-artists from different eras and geographical locations proportionate key facial regions in their portrayal of Buddha artworks. 

We propose to detect and use these proportions for the analysis and differentiation of Buddha designs from different eras and locations around the world. Fig.~\ref{fig:tibetlines} depicts the chosen iconometric proportional measurements of different facial regions that is used in our analysis. The idea is to use automatic landmark detection, so we may infer the iconometry lines from any Buddha face in the dataset. Based on these lines, we can identify and normalize the proportions of each key region of the Buddha faces and compare them together and against the canons.

\subsection{Guidelines and Proportions}

The guidelines are given for a front facing Buddha statue, but not all pictures are perfectly facing front the camera source point. Finding 3D facial landmarks allows for affine spatial transformation, and for normalizing the statue pose before searching for the iconometric guidelines. 

Moreover, we wish to locate the guidelines with relation to important facial points. To do so, we first employ facial landmark detection on the historical Buddha model, and find correspondences between the lines and the model (as detailed in Table~\ref{tab:lineconnections} and illustrated in Fig.~\ref{fig:tibetlines}-c). Because the landmark points are defined in a 3-dimensional space, the correspondences are defined on the 2D front-facing orthogonal projection of the landmarks. We employ the Position Map Regression Network (PRN)~\cite{feng2018joint} which identifies 68 3D facial landmarks in faces. Table~\ref{tab:lineconnections} defines the proportional guidelines that can be drawn from any given 68 facial landmark points (refer to the Fig.~\ref{fig:linelabels} for reference to the point numbers).

\begin{table}[b]
	\caption{The proportional guidelines can be drawn from any given 68 facial landmark points as shown in Fig.~\ref{fig:linelabels}.}\label{table1}
	\centering
	\begin{tabular}{c c c}
		\hline
		\small
		\centering
		\hspace{-1mm} Line
		&\hspace{-1mm} \small Description \hspace{-1mm}
		& \hspace{-1mm}\small  Point Connections \hspace{-1mm}
		\\
		\hline
		
		\footnotesize
		\centering L1
		& \footnotesize Eyebrow Line
		& \footnotesize Mean of (19,21) to Mean of (24,26)
		
		\\
		\footnotesize
		\centering L2
		& \footnotesize Top Eye Line
		& \footnotesize Mean of (38,39) to Mean of (44,45)
		
		\\
		\footnotesize
		\centering L3
		& \footnotesize Bottom Eye Line
		& \footnotesize Mean of (41,42) to Mean of (47,48)
		
		\\
		\footnotesize
		\centering L4
		& \footnotesize Nose Sides Line
		& \footnotesize 32 to 36
		
		\\
		\footnotesize
		\centering L5
		& \footnotesize Jaw Line
		& \footnotesize 7 to 11
		
		\\
		\footnotesize
		\centering L6
		& \footnotesize Center Nose Line
		& \footnotesize Mean of (22,23) to Mean of (28,29,30,31)
		
		\\
		\footnotesize
		\centering L7
		& \footnotesize Left Face Line
		& \footnotesize Line between L1 and L5 through 2
		
		\\
		\footnotesize
		\centering L8
		& \footnotesize Left Face Line
		& \footnotesize Line between L1 and L5 through 16
		
		\\
		\hline 
	\end{tabular}
	\label{tab:lineconnections}
\end{table}

 Once the guidelines are established from the detected 68 landmark points, each key region of the Buddha face is then measured according to the proposed proportions as seen in Fig. \ref{fig:tibetlines}. For this analysis we do not make use of the inner diagonal guidelines, but we rather focus on a clear subset of six key facial regions, namely, \textit{left forehead (LH)}, \textit{right forehead (RH)}, \textit{eyelids (EL)}, \textit{eyes (E)}, \textit{nose (N)}, and \textit{lower face (LF)}. Table~\ref{tab:proportionmeasurements} details how we may derive the proportions from the lines, with their theoretical values, Fig.~\ref{fig:model68landmarks} shows the lines once the whole process is applied to the historical model. Fig.~\ref{fig:buddha68landmarks} shows the PRN-detected 68 landmark points on a Buddha face and its 2D frontal orthographic projection is presented in Fig.~\ref{fig:buddha2D68frontal}. Results on statues are shown in Fig.~\ref{fig:lineschina}-i.\

\subsection{Analysis}

Given our dataset, we apply the above described iconometric proportions for the three main categories of statues. Given that we may have multiple pictures for each statue and that the landmark detection may fail on some pictures, we obtain 179 measurements for statues from China, 894 proportions for Japan Heian statues, and 1994 for Japan Kamakura statues. Results are reported in Fig.~\ref{fig:allhistorical} against two baselines, the theoretical Tibetan canon baseline, and the actually measured baseline on the same Tibetan model.

Although the proportion differences might be minute, it can be observed that the Buddha designs from China, in general, have much larger noses and shorter eyelids when compared with the other two datasets, while Buddhas from the Kamakura period have their design proportions in-between the other two datasets. Eyelids tend to be slightly smaller for Kamakura designs in comparison to Heian ones.
Fig.~\ref{fig:lineschina}-i show a sample of the iconometric proportional measurement taken from each of the experimented dataset while Fig.~\ref{fig:linescombined} displays a superimposition of the three.

\begin{table}[b]
	\caption{The iconometric measurements derived from the guidelines with their theoretical values, normalized by the largest possible proportion (here the total width, LH+RH=12).}
	\centering
	
	\begin{tabular}{c c c c}
		\hline
		\small
		\centering
		\hspace{-1mm} Label
		&\hspace{-1mm} \small Description \hspace{-1mm}
		& \hspace{-1mm}\small Line/Point Connections \hspace{-1mm}
		& \hspace{-1mm}\small \vtop{\hbox{\strut Theoretical length}\hbox{\strut \hspace{3mm}(normalized)}} \hspace{-1mm}
		\\
		\hline
		
		\footnotesize
		\centering LH
		& \footnotesize Left Forehead
		& \footnotesize L1 left-point to L6 top-point
		& \footnotesize 6 (0.500)
		
		\\
		\footnotesize
		\centering RH
		& \footnotesize Left Forehead
		& \footnotesize L6 top-point to L1 right-point
		& \footnotesize 6 (0.500)
		
		\\
		\footnotesize
		\centering EL
		& \footnotesize Eyelid
		& \footnotesize L1 right-point to L2 right-point
		& \footnotesize 1 (0.083)
		
		\\
		\footnotesize
		\centering E
		& \footnotesize Eye
		& \footnotesize L2 right-point to L3 right-point
		& \footnotesize 1 (0.083)
		
		\\
		\footnotesize
		\centering N
		& \footnotesize Nose
		& \footnotesize L3 right-point to L4 right-point
		& \footnotesize 2 (0.167)
		
		\\
		\footnotesize
		\centering LF
		& \footnotesize Lower Face
		& \footnotesize L4 right-point to L5 right-point
		& \footnotesize 4 (0.333)

		\\
		\hline
	\end{tabular}
	\label{tab:proportionmeasurements}
\end{table}

One can also notice some important difference between the theoretical canons of the Tibetan model and their actual measurement in the dataset. Considering the small average distance between the observed model proportions and the different measurements on real statues, we may wonder whether this distance is an artifact due to the measurement methodology -- which is trained for human faces -- or to an actual approximation of these measures. Even in the original Tibetan model, the proportions of the nose appear to the eye larger than the one originally described.

Although the differences are not striking for the measurements themselves, they do actually differ as the timelines and locations change. This motivates us to further investigate if modern image embedding can reveal further differences among different categories of Buddha statues.

\section{Modern Embeddings}\label{sec:classification}

Since the previous method based on a historical description of facial landmarks does not give a clear cut between classes, we also explore modern types of embeddings designed for classification, namely, image embeddings that take full image for description, face embeddings trained for facial recognition, and graph embeddings purely built on the semantics of the statues.

\begin{table*}[hbt]
	\caption{F1-score with weighted average on the different classification tasks for each proposed embedding.}
	\centering
	
	\begin{tabular}{c c@{\;}c c@{\;}c c@{\;}c c@{\;}c c@{\;}c c@{\;}c c@{\;}c c@{\;}c c@{\;}c}
		\hline
		\centering
		& \multicolumn{2}{c}{\hspace{-1mm} \small T1 \hspace{-1mm}}
		& \multicolumn{2}{c}{\hspace{-1mm} \small T2 \hspace{-1mm}}
		& \multicolumn{2}{c}{\hspace{-1mm} \small T3 \hspace{-1mm}}
    	& \multicolumn{2}{c}{\hspace{-1mm} \small T4 \hspace{-1mm}}
		& 
		\multicolumn{2}{c}{\hspace{-1mm} \small T5.1 \hspace{-1mm}}
		& 
		\multicolumn{2}{c}{\hspace{-1mm} \small T5.2 \hspace{-1mm}}
        & 
		\multicolumn{2}{c}{\hspace{-1mm} \small T5.3 \hspace{-1mm}}
        & 
		\multicolumn{2}{c}{\hspace{-1mm} \small T5.4 \hspace{-1mm}}
        & 
		\multicolumn{2}{c}{\hspace{-1mm} \small T5.5 \hspace{-1mm}}
        \\
		\centering
		\hspace{-1mm} \small Method
		& \multicolumn{2}{c}{\hspace{-1mm} \small Style \hspace{-1mm}}
		& \multicolumn{2}{c}{\hspace{-1mm}\small Dimensions \hspace{-1mm}}
		& \multicolumn{2}{c}{\hspace{-1mm}\small Century \hspace{-1mm}}
    	& \multicolumn{2}{c}{\hspace{-1mm}\small Statue type \hspace{-1mm}}
		& 
		\multicolumn{2}{c}{\hspace{-1mm}\footnotesize \begin{tabular}{@{}c@{}}Base \\ material\end{tabular}   \hspace{-1mm}}
		& 
		\multicolumn{2}{c}{\hspace{-1mm}\footnotesize \begin{tabular}{@{}c@{}}Color/ \\ texture\end{tabular}  \hspace{-1mm}}
		& 
		\multicolumn{2}{c}{\hspace{-1mm}\footnotesize \begin{tabular}{@{}c@{}}Type of \\ stone\end{tabular} \hspace{-1mm}}
		& 
		\multicolumn{2}{c}{\hspace{-1mm}\footnotesize \begin{tabular}{@{}c@{}}Type of \\ wood\end{tabular} \hspace{-1mm}}
		& 
		\multicolumn{2}{c}{\hspace{-1mm}\footnotesize \begin{tabular}{@{}c@{}}Construct. \\ method\end{tabular} \hspace{-1mm}}
		\\

		& \footnotesize SVM 
		& \footnotesize NN 
		& \footnotesize SVM 
		& \footnotesize NN 
		& \footnotesize SVM 
		& \footnotesize NN 
		& \footnotesize SVM 
		& \footnotesize NN 
		& \footnotesize SVM 
		& \footnotesize NN 
		& \footnotesize SVM 
		& \footnotesize NN 
		& \footnotesize SVM 
		& \footnotesize NN 
		& \footnotesize SVM 
		& \footnotesize NN 
		& \footnotesize SVM 
		& \footnotesize NN 
		\\
		\hline
		\centering 
		\footnotesize $Iconometry$
		
        & \footnotesize 0.50	& \footnotesize \textit{0.51}	
        & \footnotesize \textit{0.50}	& \footnotesize 0.48	
        & \footnotesize \textit{0.67}	& \footnotesize 0.55	
        & \footnotesize 0.35	& \footnotesize \textit{0.68}	
        & \footnotesize 0.80	& \footnotesize \textit{0.88}	
        & \footnotesize \textit{0.34}	& \footnotesize 0.23	
        & \footnotesize 0.17	& \footnotesize \textit{0.23}	
        & \footnotesize \textit{0.84}	& \footnotesize 0.83	
        & \footnotesize 0.23	& \footnotesize \textit{0.35}

		\\
		\hline

		\centering 
		\footnotesize $VGG16_{full}$
 	& \footnotesize 0.88	& \footnotesize 0.95	
 	& \footnotesize 0.54	& \footnotesize 0.74	
 	& \footnotesize 0.52	& \footnotesize 0.73	
 	& \footnotesize 0.63	& \footnotesize 0.73
 	& \footnotesize 0.89	& \footnotesize 0.82
 	& \footnotesize 0.69	& \footnotesize 0.61	
 	& \footnotesize 0.34	& \footnotesize 0.38
 	& \footnotesize 0.89	& \footnotesize 0.86
 	& \footnotesize 0.63	& \footnotesize 0.65
		\\

		\centering 
		\footnotesize $ResNet50_{full}$
 		& \footnotesize 0.88	& \footnotesize \textbf{\textit{0.98}}	
 		& \footnotesize 0.38	& \footnotesize \textbf{\textit{0.78}}
 		& \footnotesize 0.50	& \footnotesize \textbf{\textit{0.78}}	
 		& \footnotesize 0.47	& \footnotesize \textbf{\textit{0.82}}
 		& \footnotesize \textbf{\textit{0.93}}	& \footnotesize 0.86
 		& \footnotesize \textbf{\textit{0.79}}	& \footnotesize 0.66
 		& \footnotesize \textbf{\textit{0.49}}	& \footnotesize 0.42
 		& \footnotesize \textit{0.90}	& \footnotesize 0.84	
 		& \footnotesize 0.69	& \footnotesize \textit{0.70}
		\\
		
		
		
		
		
		
		
 		
		\hline

		\centering 
		\footnotesize$VGG16_{cropped}$
 		& \footnotesize 0.83	& \footnotesize 0.92	
 		& \footnotesize 0.54	& \footnotesize 0.70	
 		& \footnotesize 0.50	& \footnotesize 0.72	
 		& \footnotesize 0.67	& \footnotesize 0.69
 		& \footnotesize 0.87	& \footnotesize 0.85	
 		& \footnotesize 0.61	& \footnotesize 0.55	
 		& \footnotesize \textit{0.46}	& \footnotesize 0.37	
 		& \footnotesize 0.87	& \footnotesize 0.86	
 		& \footnotesize 0.63	& \footnotesize 0.62
		
		\\
		\centering 
		\footnotesize$ResNet50_{cropped}$
        & \footnotesize 0.88	& \footnotesize \textit{0.96}	
        & \footnotesize 0.33	& \footnotesize \textit{0.73}	
        & \footnotesize 0.50	& \footnotesize \textit{0.75}	
        & \footnotesize 0.55	& \footnotesize \textit{0.74}	
        & \footnotesize \textit{0.90}	& \footnotesize 0.89	
        & \footnotesize \textit{0.74}	& \footnotesize 0.67	
        & \footnotesize 0.45	& \footnotesize 0.39	
        & \footnotesize \textbf{\textit{0.91}}	& \footnotesize 0.86	
        & \footnotesize 0.72	& \footnotesize \textbf{\textit{0.74}}
        		\\
		\hline
		
		\centering 
		\footnotesize$VGG16_{vggface2}$
 		& \footnotesize 0.72	& \footnotesize \textit{0.89}
 		& \footnotesize 0.54	& \footnotesize 0.73
 		& \footnotesize 0.44	& \footnotesize \textit{0.70}	
 		& \footnotesize 0.67	& \footnotesize 0.71	
 		& \footnotesize 0.86	& \footnotesize 0.74	
 		& \footnotesize \textit{0.69}	& \footnotesize 0.61	
 		& \footnotesize \textit{0.43}	& \footnotesize 0.35	
 		& \footnotesize \textit{0.88}	& \footnotesize 0.85	
 		& \footnotesize \textit{0.67}	& \footnotesize 0.65
		
		\\
		\centering 
		\footnotesize$ResNet50_{vggface2}$
        & \footnotesize 0.72	& \footnotesize 0.88	
        & \footnotesize 0.54	& \footnotesize \textit{0.74}
        & \footnotesize 0.44	& \footnotesize 0.69	
        & \footnotesize 0.67	& \footnotesize \textit{0.72}	
        & \footnotesize 0.86	& \footnotesize \textit{0.87}	
        & \footnotesize \textit{0.69}	& \footnotesize 0.64	
        & \footnotesize \textit{0.43}	& \footnotesize 0.34
        & \footnotesize \textit{0.88}	& \footnotesize 0.84	
        & \footnotesize \textit{0.67}	& \footnotesize 0.65

		
		
		
		
		
		
		
 		
		\\
		\hline
		\centering 
		\footnotesize $Node2Vec_{KG}$
        & \footnotesize 0.92	& \footnotesize \textit{0.93}
        & \footnotesize  --	& \footnotesize  --	
        & \footnotesize 0.71	& \footnotesize \textit{0.74}
        & \footnotesize  --	& \footnotesize  --
        & \footnotesize  --	& \footnotesize  --	
        & \footnotesize  --	& \footnotesize  --	
        & \footnotesize  --	& \footnotesize  --
        & \footnotesize  --	& \footnotesize  --	
        & \footnotesize  --	& \footnotesize  --	
		\\

		\centering 
		\footnotesize $Node2Vec_{KG_{time}}$
        & \footnotesize \textbf{\textit{0.98}}	& \footnotesize \textbf{\textit{0.98}}	
        & \footnotesize  --	& \footnotesize  --	
        & \footnotesize  --	& \footnotesize  --	
        & \footnotesize  --	& \footnotesize  --	
        & \footnotesize  --	& \footnotesize  --	
        & \footnotesize  --	& \footnotesize  --	
        & \footnotesize  --	& \footnotesize  --	
        & \footnotesize  --	& \footnotesize  --
        & \footnotesize  --	& \footnotesize  --	
        
		\\
		\hline

	\end{tabular}
	\label{tab:T1T5}
\end{table*}

\subsection{Classification Tasks}

Our initial research question is a \textbf{style classification (T1)}, \textit{i.e.} the comparison of three different styles: \textit{China}, \textit{Kamakura period}, and \textit{Heian period}. Given the rich dataset we have been offered to explore, we also approach four additional classification tasks.

We  conduct a \textbf{statue type classification (T2)} which guesses the type of Buddha represented, and \textbf{dimension classification (T3)} which classifies the dimension of a statue across the three classes determined in Sec.~\ref{sec:data}.

We continue with a \textbf{century classification (T4)}, given the temporal alignment of our statues, each could be assigned to a different century (we are covering a total of nine centuries in our dataset).

We conclude with the \textbf{material classifications (T5)}, which comprises: \textit{base material (T5.1)}, \textit{color/texture (T5.2)}, \textit{type of stone (T5.3)}, \textit{type of wood (T5.4)}, and \textit{construction method (T5.5)}. Note that all material classifications except for task \textit{T5.5} are actually multi-label classification tasks, indeed a statue can combine different materials, colors, \textit{etc.} Only the construction method is unique, thus single label classification.

To evaluate classification across each of these tasks, we limit our dataset to the 1393 annotated and cleaned statues, covering a total of 3315 images. To compare the different methods on the same dataset, we further limit our evaluation to the 2508 pictures with a detectable face as searched during Sec.~\ref{sec:iconometry}, using PRN~\cite{feng2018joint}. Due to the limited size of the dataset, we train our classifiers using a 5-fold cross-validation.

\subsection{Image Embeddings}

To describe our Buddha statue 2D pictures, we propose to study existing neural network architectures which already have proven great success in many classification tasks, namely $VGG16$~\cite{simonyan2014very} and $ResNet50$~\cite{he2016deep}. 

For the classification of Buddha statues from the global aspect of their image, we use each of these networks with their standard pre-trained weights (from ImageNet~\cite{deng2009imagenet}).

To study the classification performances of statues with regards to their face, we first restrain the face region using PRN~\cite{feng2018joint}. To compare the relevance of the facial region for classification, we evaluate against two datasets. The first one evaluates ImageNet-trained embeddings on the full images (referred to as $VGG16_{full}$ and $ResNet50_{full}$), the second one evaluates the same features, but only on the cropped region of the face ($VGG16_{cropped}$ and $Res$\-$Net50_{cropped}$). 
In addition, each of the networks is also fine-tuned using VGGFace2~\cite{VGGFace2}, a large-scale dataset designed for the face recognition task (on cropped faces), herafter $VGG16_{vggface2}$ and $ResNet50_{vggface2}$.

Whichever the method described above, the size of the resulting embedding space is of 2048 dimensions.

\subsection{Semantic Embedding}

Given the rich data we are provided, and inspired by the work of Garcia \textit{et al.}~\cite{garcia2019context}, we may also explore semantic embedding in the form of an \textit{artistic knowledge graph}.

Instead of traditional homophily relationships, our \textit{artistic knowledge graph} $KG=(V,E)$ is composed of multiple types of node: first of all, each statue picture is a node (\textit{e.g.} the Great Buddha of Kamakura). Then, each value of each family of attributes also has a node, connected to the nodes of the statues they qualify (for example, the \textit{Great Buddha of Kamakura} node will be connected to the \textit{Bronze} node).

\begin{figure}[t]
	\centering
\includegraphics[width=0.8\linewidth]{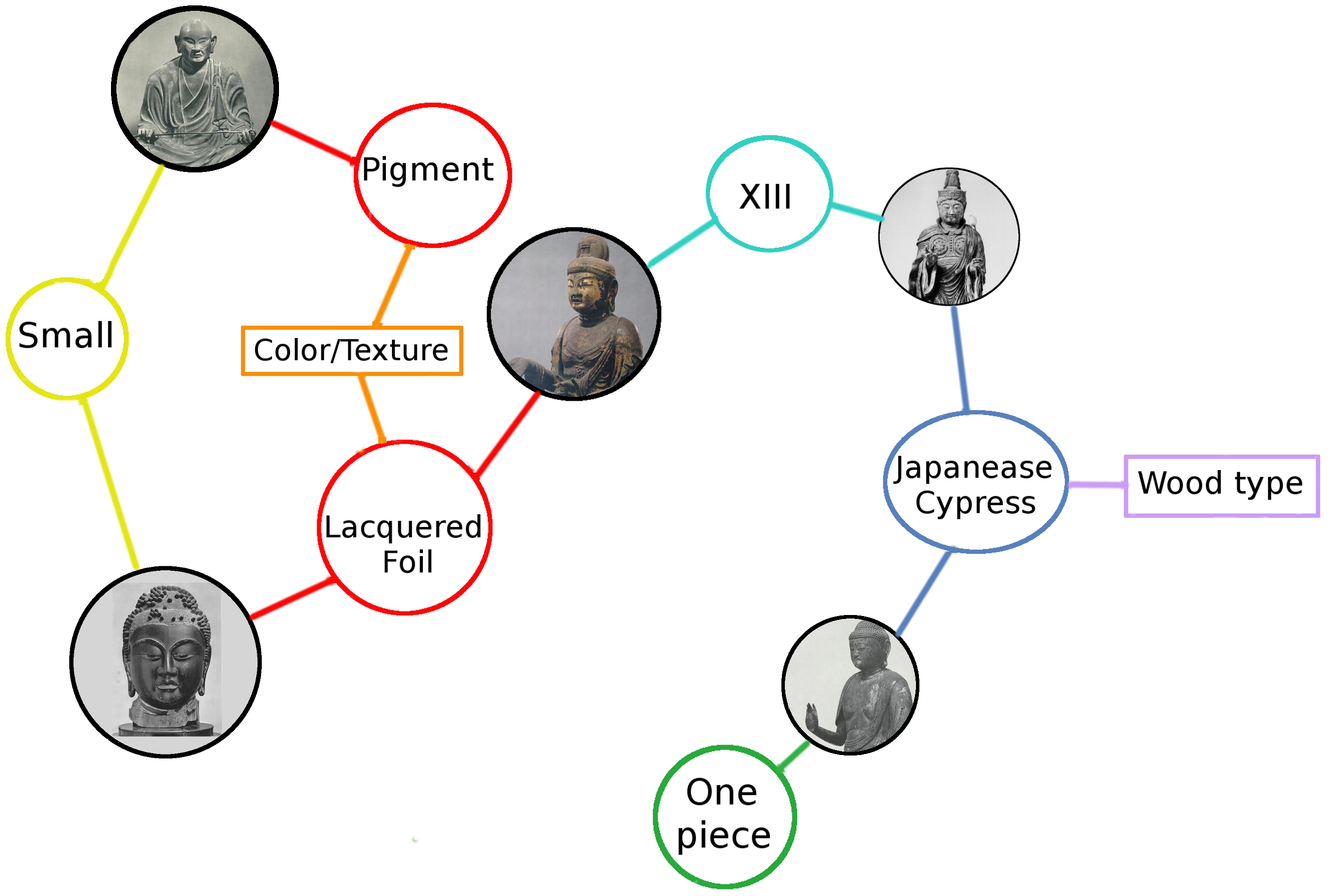}
	\caption{An example of \textit{artistic knowledge graph}. Each node corresponds to either a statue or an attribute, whereas edges correspond to existing interconnections.}
	\label{fig:KG_example}
\end{figure}

From the metadata provided, we construct two knowledge graphs. A first knowledge graph $KG$ only uses the following families of attributes: \textit{Dimensions}, \textit{Materials}, and \textit{Statue type}. Because we are curious in testing the impact of time as a determinant of style, we also add the \textit{Century} attributes in a more complete graph $KG_{time}$. In total, the resulting $KG$ presents 3389 nodes and 16756 edges, and $KG_{time}$ presents 3401 nodes and 20120 edges. An illustrative representation of our \textit{artistic knowledge graph} is shown in Fig.~\ref{fig:KG_example}.

Considering the sparsity of all our data, due to noisy and/or missing values, this graph definition suits very well our case. However, because we use category labels during the knowledge graphs construction, it limits us to evaluate only task T1 and T3 for $KG$, and T1 only for $KG_{time}$. To measure node embeddings in this graph, we use node2vec~\cite{node2vec}, which assigns a 128-dimensional representation of a node as a function of its neighborhood at a geodesic distance of 2. This should reflect statue homophily very well since statues nodes may be reached between them from a geodesic distance of 2.

\subsection{Evaluation}

We use two types of classifiers for each task. 

Given the small amount and imbalanced data we have for each different classification, we first train a classical Support Vector Machine (SVM) classifier~\cite{cortes1995support}. To improve the quality of the classifier given imbalanced data, we adjust the following parameters: $\gamma = 1/|M|$ ($|M|$, the number of classes), $penalty=1$, \textit{linear} kernel, and adjusted class weights $w_m = N/k.n_m$ (inversely proportional to class frequency in the input data: for a class $m$ among $k$ classes, having $n_m$ observations among $N$ observations).

We additionally train a Neural Network classifier (NN), in form of a fully connected layer followed by softmax activation with categorical crossentropy loss $\mathcal{L}(y, \hat{y})$ if only one category is applicable:
$$
\mathcal{L}(y, \hat{y}) = - \sum_{j=0}^M\sum_{i=0}^N(y_{ij}*\log(\hat{y}_{ij}))
$$
with $M$ categories. Otherwise, we use a binary crossentropy $H_p(q)$ for multi-label classification, as follows: 
$$H_p(q)=\frac{1}{N}\sum_{i=1}^Ny_i\log(p(y_i))+(1-y_i)\log(1-p(y_i))$$
With $y$ the embedding vector, $N$ is the training set.  
Both cases use Adam optimizer, and the size of the output layer is then matched to the number of possible classes. 

For each of the task we report the weighted average, more adapted for classification with unbalanced labels, of precision and recall under the form of F1-score (precision and recall values are very comparable across all our classifiers, so F1-score works very well in our case). Classification results are presented in Table~\ref{tab:T1T5}.

\begin{figure*}[h]
	\centering
        \begin{subfigure}[b]{0.32\textwidth}
			\centering
			\includegraphics[width=\textwidth]{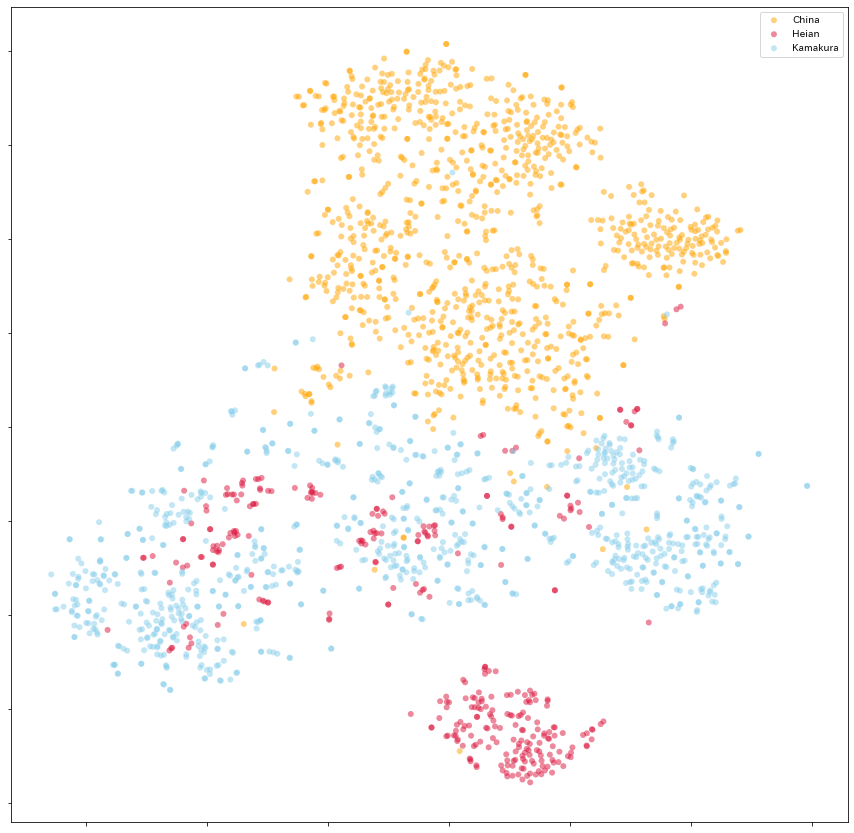}
			\caption{$ResNet50_{full}$}
			\label{fig:resnet50}
		\end{subfigure}
        \begin{subfigure}[b]{0.32\textwidth}
			\centering			\includegraphics[width=\textwidth]{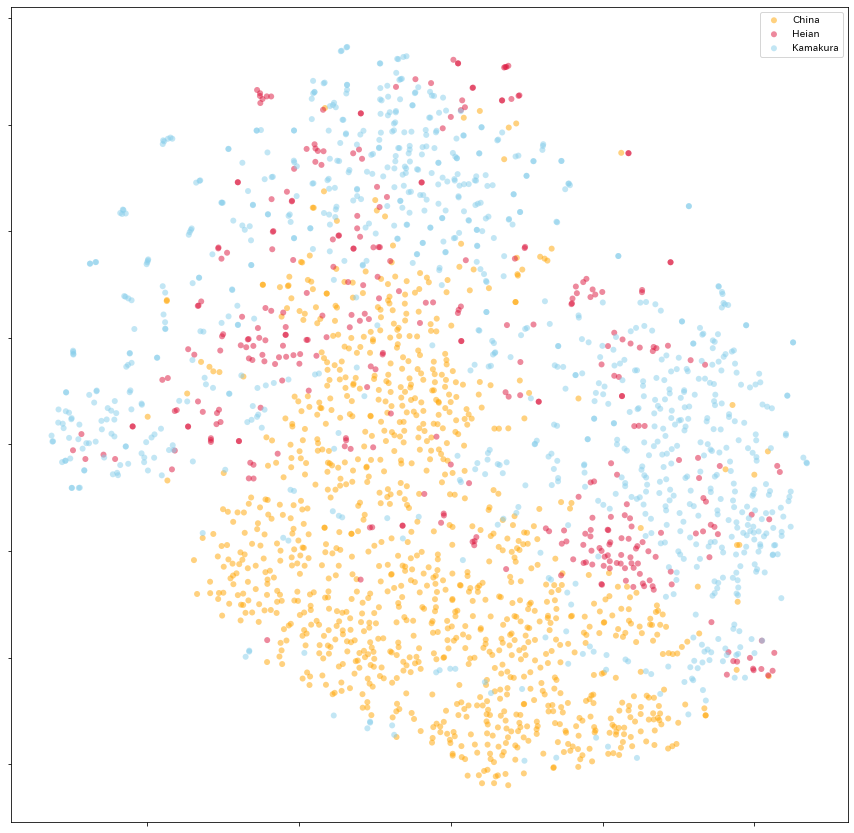}			\caption{$ResNet50_{cropped}$}
			\label{fig:resnet50cropped}
		\end{subfigure}
        \begin{subfigure}[b]{0.32\textwidth}
			\centering
			\includegraphics[width=\textwidth]{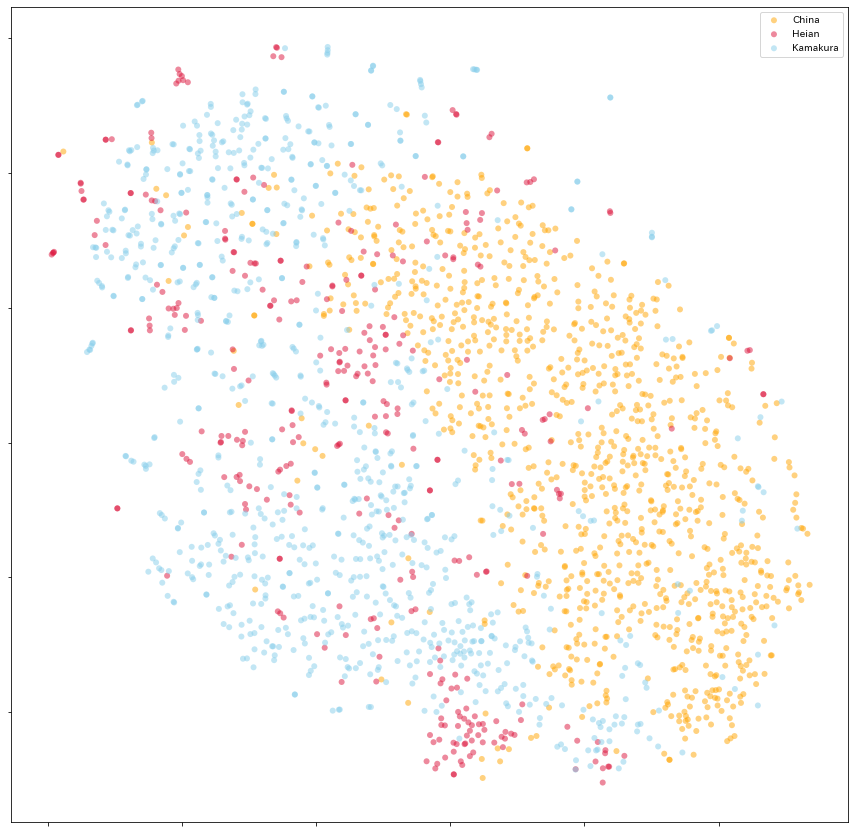}
			\caption{$ResNet50_{vggface2}$}
			\label{fig:resnetface}
		\end{subfigure}
        \begin{subfigure}[b]{0.32\textwidth}
			\centering
			\includegraphics[width=\textwidth]{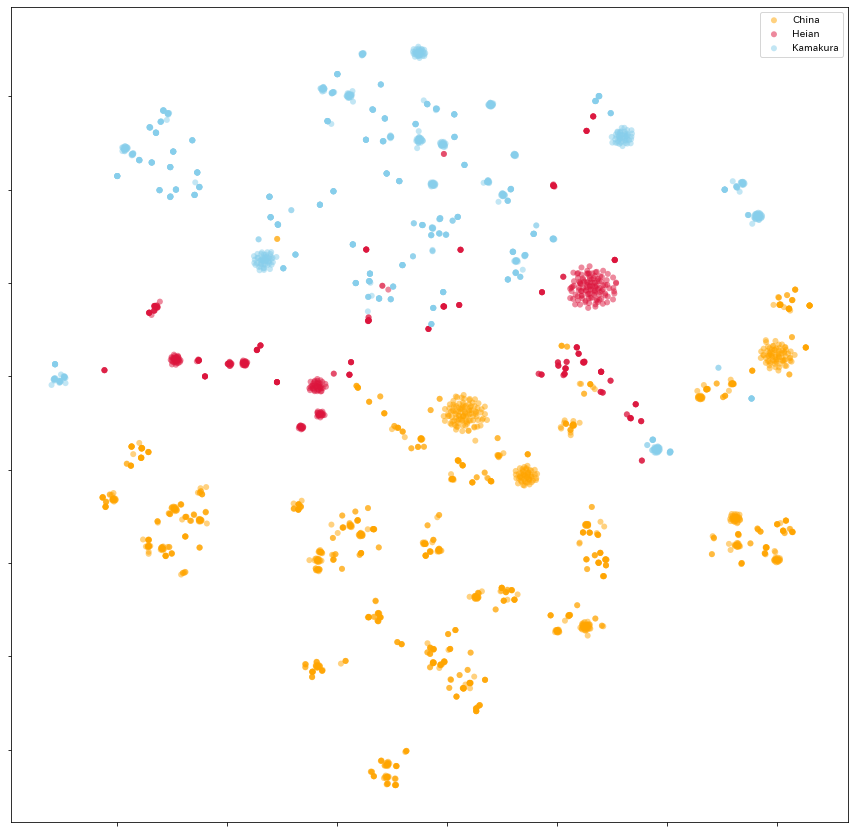}
			\caption{$node2vec_{KG_{time}}$}
			\label{fig:node2vec}
		\end{subfigure}
        \begin{subfigure}[b]{0.32\textwidth}
			\centering
			\includegraphics[width=\textwidth]{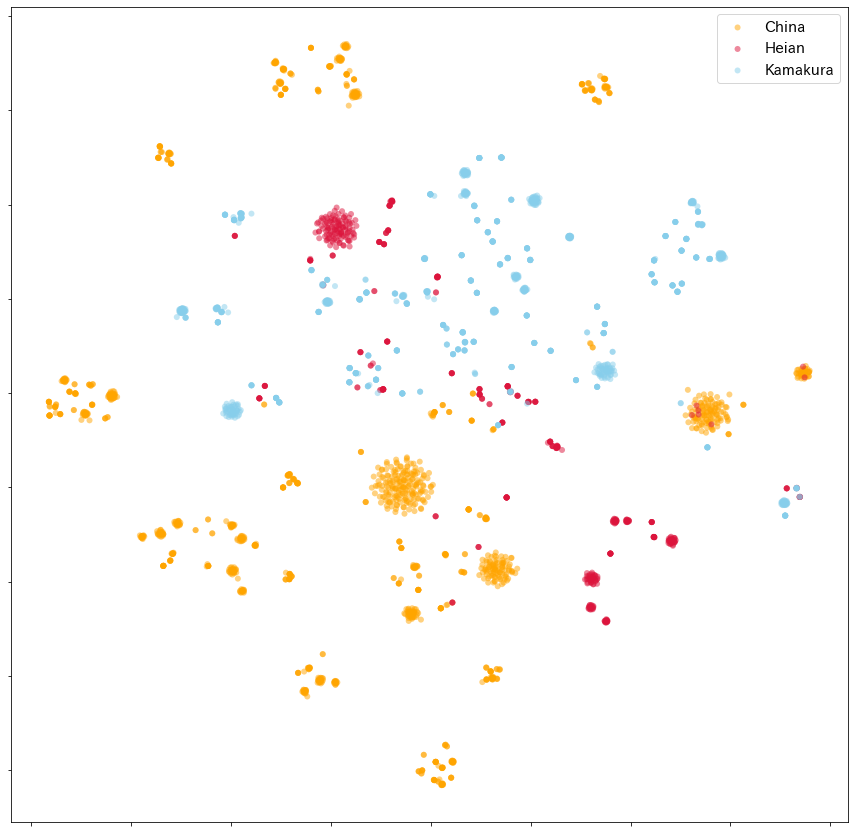}
			\caption{$node2vec_{KG}$}
			\label{fig:node2vec_emb}
		\end{subfigure}		
        \begin{subfigure}[b]{0.32\textwidth}
			\centering
			\includegraphics[width=\textwidth]{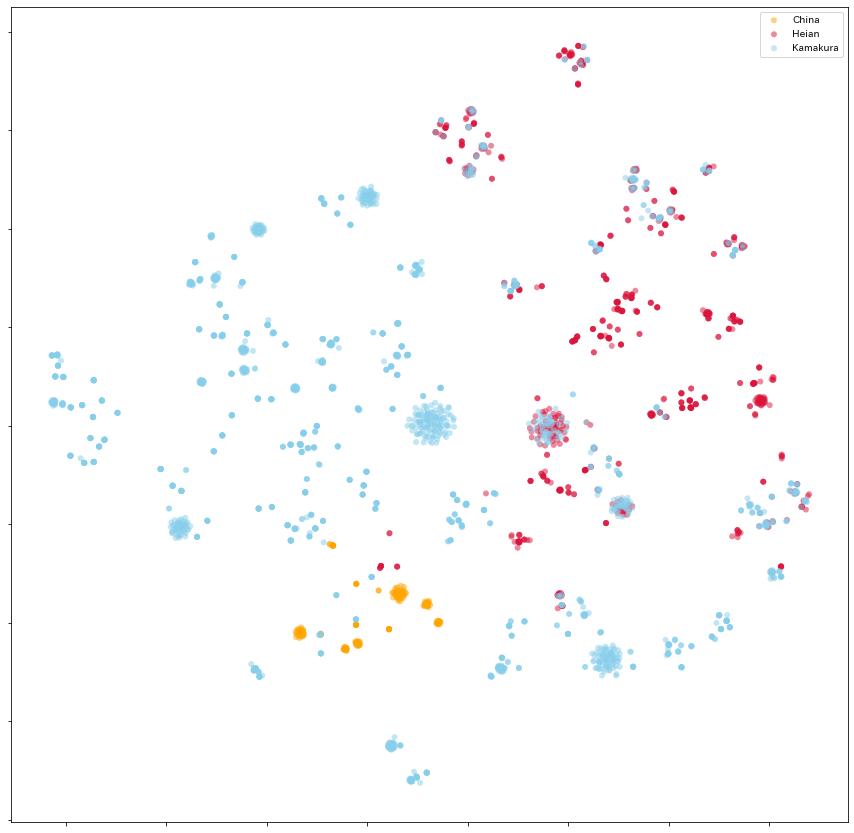}
			\caption{$Iconometry$}
			\label{fig:iconmetry_emb}
		\end{subfigure}
	\caption{Comparison of all embeddings through a 2D projection with tSNE~\cite{maaten2008visualizing}, colored with the three styles China (orange), Heian (red), and Kamakura (blue).}
	\label{fig:alltsne}
\end{figure*}

\section{Discussion}

 The first point we may notice from the classification results is that using iconometry does not perform very well in comparison to neural networks. In average, we obtain with iconometry a precision score of 0.49 for a recall score of 0.67, whereas scores are very similar for all the other methods. 
 Deep learning based methods perform better, but not equally on all tasks. 
 Style (T1), base material (T5.1) and type of wood (T5.4) are the tasks that are the best classified. Type of stone (T5.3) is by far the most difficult to classify, probably due to the imbalance of its labels. Iconometry is significantly worse than the other methods in guessing the construction method (T5.5) and the color/texture (T5.2). The neural network classifier (NN) usually perform better than SVM, except in the multilabel classification tasks (T5.1--T5.4). It suggests that those classes have a more linear distribution. 
 
  VGGFace2 ~\cite{VGGFace2} trained methods perform a little worse than their counterpart trained with ImageNet~\cite{deng2009imagenet} on the cropped faces, which in turn perform slightly worse than the full images. This may happen because VGGFace2 dataset takes into account variations between ages, contrarily to the shape of Buddha faces which are more rigid (from the point of view of the construction guidelines). It might also suggest that faces of Buddha statues differ fundamentally from a standard human face, which makes the transfer learning not so effective from the networks pretrained in VGGFace2. This proposition agrees with the fact that iconometry did not perform well. The differences are not significant though, giving us great hope for fine tuning models directly from Buddha faces.
  
 In addition, $ResNet50$~\cite{he2016deep} appears to show the best results overall. Remarkably, when we focus only on the face region, $ResNet50$ performs even better than the others when classifying \textbf{type of wood} and \textbf{construction method},  which encourages the idea of using face region for a good discriminator, specially for classification related to the material. 

The semantic embeddings based on \textit{artistic knowledge graph} perform as well as the best of image-based embeddings for style classification (T1), a result consistent with Garcia \textit{et al.}'s observations~\cite{garcia2019context}. This is probably due to the contextual information carried by the graph. 
However, if the century is not present in the KG, $ResNet50$ still shows better results than $node2vec$.
We can additionally underline that temporal information is a good predictor of \textbf{style}, since the classification performance is slightly improved after adding the \textbf{centuries} information in the knowledge graph.

We may further investigate the space defined by those embeddings as illustrated in Fig.\ref{fig:alltsne}. It is interesting to see the similarities in the space between $node2vec$ and the iconometry used as embeddings. However, their classification performances are very different. The iconometry embeddings do not look like to well separate the three styles, but there seems to be quite notable clusters forming that should be interesting to investigate further. The advantage of iconometry over other embeddings is its explainability. Integrating time into the $KG$ clearly shows a better separatibility of the three styles.

By looking at the spread of the face-based $ResNet50_{cropped}$ $ResNet50_{face}$ embeddings, we may also notice that the different classes are much more diffused than $ResNet50_{full}$. The face region is very specialized in the space of all shapes. Although the $vggface2$ embeddings are trained for a different task, facial recognition, \textit{i.e.} to identify similar faces with quite some variability, we do not see a clear difference between the separation of style from the face regions. To show the effectiveness of facial analysis against whole picture analysis, we will need to proceed with further experiments, including increasing the variety of Buddha statues in our dataset in order to train specific models designed for Buddha faces.

\section{Conclusion}

We have presented a method for acquisition of iconometric guidelines in Buddha faces and use them for different classification tasks. We have compared them with different modern embeddings, which have demonstrated much higher classification performances. Still there is one advantage of the iconometric guidelines from their simplicity and ease of understanding. To further understand what makes a style, we would like to investigate visualization and parameter regressions in future works, and identify salient areas that are specific to a class.

We have presented one straightforward method for the identification of iconometric landmarks in Buddha statue, but many statues did not show good enough landmarks to be measured at all. We could extend our landmark analysis, and boost the discrimination power of landmarks by designing a specific landmark detector for Buddha statues. 

Scanning books is definitely more scalable today than 3D-captures of the statues. However, with the high results of the deep-learning methods for style classification, we could question how influential was the data acquisition method on the classification. Each book paper may have a slightly different grain that deep neural networks may have captured. Nonetheless, the different classification tasks are relatively independent from the book source while still showing quite high results. One of our goals is to continue develop this dataset and multiply our data sources, so it would further diminish the influence of data acquisition over analysis.

\textbf{Acknowledgement:} This work was supported by JSPS KAKENHI Grant Number 18H03571.


\bibliographystyle{ACM-Reference-Format}
\bibliography{bibliography}


\begin{thebibliography}{50}


\ifx \showCODEN    \undefined \def \showCODEN     #1{\unskip}     \fi
\ifx \showDOI      \undefined \def \showDOI       #1{#1}\fi
\ifx \showISBNx    \undefined \def \showISBNx     #1{\unskip}     \fi
\ifx \showISBNxiii \undefined \def \showISBNxiii  #1{\unskip}     \fi
\ifx \showISSN     \undefined \def \showISSN      #1{\unskip}     \fi
\ifx \showLCCN     \undefined \def \showLCCN      #1{\unskip}     \fi
\ifx \shownote     \undefined \def \shownote      #1{#1}          \fi
\ifx \showarticletitle \undefined \def \showarticletitle #1{#1}   \fi
\ifx \showURL      \undefined \def \showURL       {\relax}        \fi
\providecommand\bibfield[2]{#2}
\providecommand\bibinfo[2]{#2}
\providecommand\natexlab[1]{#1}
\providecommand\showeprint[2][]{arXiv:#2}

\bibitem[\protect\citeauthoryear{Bar, Levy, and Wolf}{Bar
  et~al\mbox{.}}{2014}]%
        {Bar2014ClassificationOA}
\bibfield{author}{\bibinfo{person}{Yaniv Bar}, \bibinfo{person}{Noga Levy},
  {and} \bibinfo{person}{Lior Wolf}.} \bibinfo{year}{2014}\natexlab{}.
\newblock \showarticletitle{Classification of Artistic Styles Using Binarized
  Features Derived from a Deep Neural Network}. In
  \bibinfo{booktitle}{\emph{ECCV Workshops}}.
\newblock


\bibitem[\protect\citeauthoryear{Bevan, Li, Martinon-Torres, Green, Xia, Zhao,
  Zhao, Ma, Cao, and Rehren}{Bevan et~al\mbox{.}}{2014}]%
        {bevan2014computer}
\bibfield{author}{\bibinfo{person}{Andrew Bevan}, \bibinfo{person}{Xiuzhen Li},
  \bibinfo{person}{Marcos Martinon-Torres}, \bibinfo{person}{Susan Green},
  \bibinfo{person}{Yin Xia}, \bibinfo{person}{Kun Zhao}, \bibinfo{person}{Zhen
  Zhao}, \bibinfo{person}{Shengtao Ma}, \bibinfo{person}{Wei Cao}, {and}
  \bibinfo{person}{Thilo Rehren}.} \bibinfo{year}{2014}\natexlab{}.
\newblock \showarticletitle{Computer vision, archaeological classification and
  China's terracotta warriors}.
\newblock \bibinfo{journal}{\emph{Journal of Archaeological Science}}
  \bibinfo{volume}{49} (\bibinfo{year}{2014}), \bibinfo{pages}{249--254}.
\newblock


\bibitem[\protect\citeauthoryear{Bhaumik, Samaddar, and Samaddar}{Bhaumik
  et~al\mbox{.}}{2018}]%
        {bhaumik2018recognition}
\bibfield{author}{\bibinfo{person}{Gopa Bhaumik},
  \bibinfo{person}{Shefalika~Ghosh Samaddar}, {and} \bibinfo{person}{Arun~Baran
  Samaddar}.} \bibinfo{year}{2018}\natexlab{}.
\newblock \showarticletitle{Recognition Techniques in Buddhist Iconography and
  Challenges}. In \bibinfo{booktitle}{\emph{2018 International Conference on
  Advances in Computing, Communications and Informatics (ICACCI)}}. IEEE,
  \bibinfo{pages}{1285--1289}.
\newblock


\bibitem[\protect\citeauthoryear{Blessing and Wen}{Blessing and Wen}{2010}]%
        {blessing2010using}
\bibfield{author}{\bibinfo{person}{Alexander Blessing} {and}
  \bibinfo{person}{Kai Wen}.} \bibinfo{year}{2010}\natexlab{}.
\newblock \showarticletitle{Using machine learning for identification of art
  paintings}.
\newblock \bibinfo{journal}{\emph{Technical report}} (\bibinfo{year}{2010}).
\newblock


\bibitem[\protect\citeauthoryear{Cao, Shen, Xie, Parkhi, and Zisserman}{Cao
  et~al\mbox{.}}{2018}]%
        {VGGFace2}
\bibfield{author}{\bibinfo{person}{Q. Cao}, \bibinfo{person}{L. Shen},
  \bibinfo{person}{W. Xie}, \bibinfo{person}{O.~M. Parkhi}, {and}
  \bibinfo{person}{A. Zisserman}.} \bibinfo{year}{2018}\natexlab{}.
\newblock \showarticletitle{VGGFace2: A dataset for recognising faces across
  pose and age}. In \bibinfo{booktitle}{\emph{International Conference on
  Automatic Face and Gesture Recognition}}.
\newblock


\bibitem[\protect\citeauthoryear{Carneiro, da~Silva, Del~Bue, and
  Costeira}{Carneiro et~al\mbox{.}}{2012}]%
        {carneiro2012artistic}
\bibfield{author}{\bibinfo{person}{Gustavo Carneiro},
  \bibinfo{person}{Nuno~Pinho da Silva}, \bibinfo{person}{Alessio Del~Bue},
  {and} \bibinfo{person}{Jo{\~a}o~Paulo Costeira}.}
  \bibinfo{year}{2012}\natexlab{}.
\newblock \showarticletitle{Artistic image classification: An analysis on the
  printart database}. In \bibinfo{booktitle}{\emph{ECCV}}.
\newblock


\bibitem[\protect\citeauthoryear{Chu and Wu}{Chu and Wu}{2018}]%
        {chu2018image}
\bibfield{author}{\bibinfo{person}{Wei-Ta Chu} {and} \bibinfo{person}{Yi-Ling
  Wu}.} \bibinfo{year}{2018}\natexlab{}.
\newblock \showarticletitle{Image style classification based on learnt deep
  correlation features}.
\newblock \bibinfo{journal}{\emph{IEEE Transactions on Multimedia}}
  \bibinfo{volume}{20}, \bibinfo{number}{9} (\bibinfo{year}{2018}),
  \bibinfo{pages}{2491--2502}.
\newblock


\bibitem[\protect\citeauthoryear{Cortes and Vapnik}{Cortes and Vapnik}{1995}]%
        {cortes1995support}
\bibfield{author}{\bibinfo{person}{Corinna Cortes} {and}
  \bibinfo{person}{Vladimir Vapnik}.} \bibinfo{year}{1995}\natexlab{}.
\newblock \showarticletitle{Support-vector networks}.
\newblock \bibinfo{journal}{\emph{Machine learning}} \bibinfo{volume}{20},
  \bibinfo{number}{3} (\bibinfo{year}{1995}), \bibinfo{pages}{273--297}.
\newblock


\bibitem[\protect\citeauthoryear{Crowley and Zisserman}{Crowley and
  Zisserman}{2014}]%
        {crowley2014state}
\bibfield{author}{\bibinfo{person}{Elliot Crowley} {and}
  \bibinfo{person}{Andrew Zisserman}.} \bibinfo{year}{2014}\natexlab{}.
\newblock \showarticletitle{The State of the Art: Object Retrieval in Paintings
  using Discriminative Regions.}. In \bibinfo{booktitle}{\emph{BMVC}}.
\newblock


\bibitem[\protect\citeauthoryear{Crowley, Parkhi, and Zisserman}{Crowley
  et~al\mbox{.}}{2015}]%
        {crowley2015face}
\bibfield{author}{\bibinfo{person}{Elliot~J Crowley}, \bibinfo{person}{Omkar~M
  Parkhi}, {and} \bibinfo{person}{Andrew Zisserman}.}
  \bibinfo{year}{2015}\natexlab{}.
\newblock \showarticletitle{Face Painting: querying art with photos.}. In
  \bibinfo{booktitle}{\emph{BMVC}}. \bibinfo{pages}{65--1}.
\newblock


\bibitem[\protect\citeauthoryear{Crowley and Zisserman}{Crowley and
  Zisserman}{2016}]%
        {crowley2016art}
\bibfield{author}{\bibinfo{person}{Elliot~J Crowley} {and}
  \bibinfo{person}{Andrew Zisserman}.} \bibinfo{year}{2016}\natexlab{}.
\newblock \showarticletitle{The art of detection}. In
  \bibinfo{booktitle}{\emph{ECCV}}. Springer.
\newblock


\bibitem[\protect\citeauthoryear{Deng, Dong, Socher, Li, Li, and Fei-Fei}{Deng
  et~al\mbox{.}}{2009}]%
        {deng2009imagenet}
\bibfield{author}{\bibinfo{person}{Jia Deng}, \bibinfo{person}{Wei Dong},
  \bibinfo{person}{Richard Socher}, \bibinfo{person}{Li-Jia Li},
  \bibinfo{person}{Kai Li}, {and} \bibinfo{person}{Li Fei-Fei}.}
  \bibinfo{year}{2009}\natexlab{}.
\newblock \showarticletitle{Imagenet: A large-scale hierarchical image
  database}. In \bibinfo{booktitle}{\emph{2009 IEEE conference on computer
  vision and pattern recognition}}. IEEE, \bibinfo{pages}{248--255}.
\newblock


\bibitem[\protect\citeauthoryear{Elgammal, Mazzone, Liu, Kim, and
  Elhoseiny}{Elgammal et~al\mbox{.}}{2018}]%
        {elgammal2018shape}
\bibfield{author}{\bibinfo{person}{Ahmed Elgammal}, \bibinfo{person}{Marian
  Mazzone}, \bibinfo{person}{Bingchen Liu}, \bibinfo{person}{Diana Kim}, {and}
  \bibinfo{person}{Mohamed Elhoseiny}.} \bibinfo{year}{2018}\natexlab{}.
\newblock \showarticletitle{The Shape of Art History in the Eyes of the
  Machine}.
\newblock \bibinfo{journal}{\emph{arXiv preprint arXiv:1801.07729}}
  (\bibinfo{year}{2018}).
\newblock


\bibitem[\protect\citeauthoryear{Elgammal and Saleh}{Elgammal and
  Saleh}{2015}]%
        {elgammal2015quantifying}
\bibfield{author}{\bibinfo{person}{Ahmed Elgammal} {and} \bibinfo{person}{Babak
  Saleh}.} \bibinfo{year}{2015}\natexlab{}.
\newblock \showarticletitle{Quantifying creativity in art networks}.
\newblock \bibinfo{journal}{\emph{arXiv preprint arXiv:1506.00711}}
  (\bibinfo{year}{2015}).
\newblock


\bibitem[\protect\citeauthoryear{Feng, Wu, Shao, Wang, and Zhou}{Feng
  et~al\mbox{.}}{2018}]%
        {feng2018joint}
\bibfield{author}{\bibinfo{person}{Yao Feng}, \bibinfo{person}{Fan Wu},
  \bibinfo{person}{Xiaohu Shao}, \bibinfo{person}{Yanfeng Wang}, {and}
  \bibinfo{person}{Xi Zhou}.} \bibinfo{year}{2018}\natexlab{}.
\newblock \showarticletitle{Joint 3d face reconstruction and dense alignment
  with position map regression network}. In
  \bibinfo{booktitle}{\emph{Proceedings of the European Conference on Computer
  Vision (ECCV)}}. \bibinfo{pages}{534--551}.
\newblock


\bibitem[\protect\citeauthoryear{Garcia, Renoust, and Nakashima}{Garcia
  et~al\mbox{.}}{2019}]%
        {garcia2019context}
\bibfield{author}{\bibinfo{person}{Noa Garcia}, \bibinfo{person}{Benjamin
  Renoust}, {and} \bibinfo{person}{Yuta Nakashima}.}
  \bibinfo{year}{2019}\natexlab{}.
\newblock \showarticletitle{Context-Aware Embeddings for Automatic Art
  Analysis}. In \bibinfo{booktitle}{\emph{Proceedings of the 2019 on
  International Conference on Multimedia Retrieval}}. ACM,
  \bibinfo{pages}{25--33}.
\newblock


\bibitem[\protect\citeauthoryear{Garcia and Vogiatzis}{Garcia and
  Vogiatzis}{2018}]%
        {Garcia2018How}
\bibfield{author}{\bibinfo{person}{Noa Garcia} {and} \bibinfo{person}{George
  Vogiatzis}.} \bibinfo{year}{2018}\natexlab{}.
\newblock \showarticletitle{How to Read Paintings: Semantic Art Understanding
  with Multi-Modal Retrieval}. In \bibinfo{booktitle}{\emph{EECV Workshops}}.
\newblock


\bibitem[\protect\citeauthoryear{Grover and Leskovec}{Grover and
  Leskovec}{2016}]%
        {node2vec}
\bibfield{author}{\bibinfo{person}{Aditya Grover} {and} \bibinfo{person}{Jure
  Leskovec}.} \bibinfo{year}{2016}\natexlab{}.
\newblock \showarticletitle{Node2Vec: Scalable Feature Learning for Networks}.
  In \bibinfo{booktitle}{\emph{Proceedings of the 22Nd ACM SIGKDD International
  Conference on Knowledge Discovery and Data Mining}}
  \emph{(\bibinfo{series}{KDD '16})}. \bibinfo{publisher}{ACM},
  \bibinfo{address}{New York, NY, USA}, \bibinfo{pages}{855--864}.
\newblock
\showISBNx{978-1-4503-4232-2}
\urldef\tempurl%
\url{https://doi.org/10.1145/2939672.2939754}
\showDOI{\tempurl}


\bibitem[\protect\citeauthoryear{He, Zhang, Ren, and Sun}{He
  et~al\mbox{.}}{2016}]%
        {he2016deep}
\bibfield{author}{\bibinfo{person}{Kaiming He}, \bibinfo{person}{Xiangyu
  Zhang}, \bibinfo{person}{Shaoqing Ren}, {and} \bibinfo{person}{Jian Sun}.}
  \bibinfo{year}{2016}\natexlab{}.
\newblock \showarticletitle{Deep residual learning for image recognition}. In
  \bibinfo{booktitle}{\emph{Proceedings of the IEEE conference on computer
  vision and pattern recognition}}. \bibinfo{pages}{770--778}.
\newblock


\bibitem[\protect\citeauthoryear{Ikeuchi, Oishi, Takamatsu, Sagawa, Nakazawa,
  Kurazume, Nishino, Kamakura, and Okamoto}{Ikeuchi et~al\mbox{.}}{2007}]%
        {ikeuchi2007great}
\bibfield{author}{\bibinfo{person}{Katsushi Ikeuchi}, \bibinfo{person}{Takeshi
  Oishi}, \bibinfo{person}{Jun Takamatsu}, \bibinfo{person}{Ryusuke Sagawa},
  \bibinfo{person}{Atsushi Nakazawa}, \bibinfo{person}{Ryo Kurazume},
  \bibinfo{person}{Ko Nishino}, \bibinfo{person}{Mawo Kamakura}, {and}
  \bibinfo{person}{Yasuhide Okamoto}.} \bibinfo{year}{2007}\natexlab{}.
\newblock \showarticletitle{The great buddha project: Digitally archiving,
  restoring, and analyzing cultural heritage objects}.
\newblock \bibinfo{journal}{\emph{International Journal of Computer Vision}}
  \bibinfo{volume}{75}, \bibinfo{number}{1} (\bibinfo{year}{2007}),
  \bibinfo{pages}{189--208}.
\newblock


\bibitem[\protect\citeauthoryear{Johnson, Hendriks, Berezhnoy, Brevdo, Hughes,
  Daubechies, Li, Postma, and Wang}{Johnson et~al\mbox{.}}{2008}]%
        {johnson2008image}
\bibfield{author}{\bibinfo{person}{C~Richard Johnson}, \bibinfo{person}{Ella
  Hendriks}, \bibinfo{person}{Igor~J Berezhnoy}, \bibinfo{person}{Eugene
  Brevdo}, \bibinfo{person}{Shannon~M Hughes}, \bibinfo{person}{Ingrid
  Daubechies}, \bibinfo{person}{Jia Li}, \bibinfo{person}{Eric Postma}, {and}
  \bibinfo{person}{James~Z Wang}.} \bibinfo{year}{2008}\natexlab{}.
\newblock \showarticletitle{Image processing for artist identification}.
\newblock \bibinfo{journal}{\emph{IEEE Signal Processing Magazine}}
  \bibinfo{volume}{25}, \bibinfo{number}{4} (\bibinfo{year}{2008}).
\newblock


\bibitem[\protect\citeauthoryear{Kamakura, Oishi, Takamatsu, and
  Ikeuchi}{Kamakura et~al\mbox{.}}{2005}]%
        {kamakura2005classification}
\bibfield{author}{\bibinfo{person}{Mawo Kamakura}, \bibinfo{person}{Takeshi
  Oishi}, \bibinfo{person}{Jun Takamatsu}, {and} \bibinfo{person}{Katsushi
  Ikeuchi}.} \bibinfo{year}{2005}\natexlab{}.
\newblock \showarticletitle{Classification of Bayon faces using 3D models}. In
  \bibinfo{booktitle}{\emph{Virtual Systems and Multimedia}}.
  \bibinfo{pages}{751--760}.
\newblock


\bibitem[\protect\citeauthoryear{Karayev, Trentacoste, Han, Agarwala, Darrell,
  Hertzmann, and Winnemoeller}{Karayev et~al\mbox{.}}{2014}]%
        {karayev2014recognizing}
\bibfield{author}{\bibinfo{person}{Sergey Karayev}, \bibinfo{person}{Matthew
  Trentacoste}, \bibinfo{person}{Helen Han}, \bibinfo{person}{Aseem Agarwala},
  \bibinfo{person}{Trevor Darrell}, \bibinfo{person}{Aaron Hertzmann}, {and}
  \bibinfo{person}{Holger Winnemoeller}.} \bibinfo{year}{2014}\natexlab{}.
\newblock \showarticletitle{Recognizing Image Style}. In
  \bibinfo{booktitle}{\emph{BMVC}}.
\newblock


\bibitem[\protect\citeauthoryear{Khan, Beigpour, Van~de Weijer, and
  Felsberg}{Khan et~al\mbox{.}}{2014}]%
        {khan2014painting}
\bibfield{author}{\bibinfo{person}{Fahad~Shahbaz Khan}, \bibinfo{person}{Shida
  Beigpour}, \bibinfo{person}{Joost Van~de Weijer}, {and}
  \bibinfo{person}{Michael Felsberg}.} \bibinfo{year}{2014}\natexlab{}.
\newblock \showarticletitle{Painting-91: a large scale database for
  computational painting categorization}.
\newblock \bibinfo{journal}{\emph{Machine vision and applications}}
  (\bibinfo{year}{2014}).
\newblock


\bibitem[\protect\citeauthoryear{Kubo and Murakami}{Kubo and Murakami}{2011}]%
        {controversy2011}
\bibfield{author}{\bibinfo{person}{Ayano Kubo} {and} \bibinfo{person}{Masakatsu
  Murakami}.} \bibinfo{year}{2011}\natexlab{}.
\newblock \showarticletitle{Dojidaibusshi tono hikaku niyoru kaikeisakuhin no
  tokucho nitsuite- yoshiki to horyo kara miru}.
\newblock \bibinfo{journal}{\emph{IPSJ SIG Computers and the Humanities (CH)}}
  \bibinfo{volume}{1} (\bibinfo{year}{2011}), \bibinfo{pages}{1--6}.
\newblock


\bibitem[\protect\citeauthoryear{Ma, Gao, Bai, Lou, Wang, Huang, and Duan}{Ma
  et~al\mbox{.}}{2017}]%
        {ma2017part}
\bibfield{author}{\bibinfo{person}{Daiqian Ma}, \bibinfo{person}{Feng Gao},
  \bibinfo{person}{Yan Bai}, \bibinfo{person}{Yihang Lou},
  \bibinfo{person}{Shiqi Wang}, \bibinfo{person}{Tiejun Huang}, {and}
  \bibinfo{person}{Ling-Yu Duan}.} \bibinfo{year}{2017}\natexlab{}.
\newblock \showarticletitle{From Part to Whole: Who is Behind the Painting?}.
  In \bibinfo{booktitle}{\emph{ACMMM}}. ACM.
\newblock


\bibitem[\protect\citeauthoryear{Maaten and Hinton}{Maaten and Hinton}{2008}]%
        {maaten2008visualizing}
\bibfield{author}{\bibinfo{person}{Laurens van~der Maaten} {and}
  \bibinfo{person}{Geoffrey Hinton}.} \bibinfo{year}{2008}\natexlab{}.
\newblock \showarticletitle{Visualizing data using t-SNE}.
\newblock \bibinfo{journal}{\emph{Journal of machine learning research}}
  \bibinfo{volume}{9}, \bibinfo{number}{Nov} (\bibinfo{year}{2008}),
  \bibinfo{pages}{2579--2605}.
\newblock


\bibitem[\protect\citeauthoryear{Mao, Cheung, and She}{Mao
  et~al\mbox{.}}{2017}]%
        {mao2017deepart}
\bibfield{author}{\bibinfo{person}{Hui Mao}, \bibinfo{person}{Ming Cheung},
  {and} \bibinfo{person}{James She}.} \bibinfo{year}{2017}\natexlab{}.
\newblock \showarticletitle{DeepArt: Learning Joint Representations of Visual
  Arts}. In \bibinfo{booktitle}{\emph{ACMMM}}.
\newblock


\bibitem[\protect\citeauthoryear{Matubara}{Matubara}{1995}]%
        {chinabook}
\bibfield{author}{\bibinfo{person}{Sabutou Matubara}.}
  \bibinfo{year}{1995}\natexlab{}.
\newblock \bibinfo{booktitle}{\emph{Chugokubukkyochokokushiron}}.
\newblock \bibinfo{publisher}{Yoshikawakobunkan}.
\newblock


\bibitem[\protect\citeauthoryear{Mensink and Van~Gemert}{Mensink and
  Van~Gemert}{2014}]%
        {mensink2014rijksmuseum}
\bibfield{author}{\bibinfo{person}{Thomas Mensink} {and} \bibinfo{person}{Jan
  Van~Gemert}.} \bibinfo{year}{2014}\natexlab{}.
\newblock \showarticletitle{The rijksmuseum challenge: Museum-centered visual
  recognition}. In \bibinfo{booktitle}{\emph{ICMR}}.
\newblock


\bibitem[\protect\citeauthoryear{Mizuno}{Mizuno}{2016}]%
        {Kamakurabook}
\bibfield{author}{\bibinfo{person}{Keizaburo Mizuno}.}
  \bibinfo{year}{2016}\natexlab{}.
\newblock \bibinfo{booktitle}{\emph{Nihonchokokushikisoshiryoshusei
  kamsakurajidai zozomeikihen}}.
\newblock \bibinfo{publisher}{Chuokoron Bijutsushuppan}.
\newblock


\bibitem[\protect\citeauthoryear{Nabata}{Nabata}{1986}]%
        {spread1986}
\bibfield{author}{\bibinfo{person}{Takashi Nabata}.}
  \bibinfo{year}{1986}\natexlab{}.
\newblock \showarticletitle{Bukkyodenrai to butsuzo no densetsu}.
\newblock \bibinfo{journal}{\emph{Otani Gakuho}} \bibinfo{volume}{65},
  \bibinfo{number}{4} (\bibinfo{year}{1986}), \bibinfo{pages}{p1--16}.
\newblock


\bibitem[\protect\citeauthoryear{{National Records of Scotland}}{{National
  Records of Scotland}}{2019}]%
        {tibetan}
\bibfield{author}{\bibinfo{person}{{National Records of Scotland}}.}
  \bibinfo{year}{2019}\natexlab{}.
\newblock \bibinfo{title}{Tibetan Buddhist Art}.
\newblock
\newblock
\urldef\tempurl%
\url{www.tibetanbuddhistart.com}
\showURL{%
\tempurl}
\newblock
\shownote{Last accessed: 2019-07-01.}


\bibitem[\protect\citeauthoryear{{Newar and Tibetan artists}}{{Newar and
  Tibetan artists}}{17  }]%
        {tibet17--}
\bibfield{author}{\bibinfo{person}{{Newar and Tibetan artists}}.}
  \bibinfo{year}{17--}\natexlab{}.
\newblock \bibinfo{booktitle}{\emph{The Tibetan Book of Proportions}}.
\newblock


\bibitem[\protect\citeauthoryear{Nishimura and Ogawa}{Nishimura and
  Ogawa}{1987}]%
        {style1987}
\bibfield{author}{\bibinfo{person}{Kocho Nishimura} {and} \bibinfo{person}{Kozo
  Ogawa}.} \bibinfo{year}{1987}\natexlab{}.
\newblock \showarticletitle{Butsuzo no miwakekata}.
\newblock  (\bibinfo{year}{1987}).
\newblock


\bibitem[\protect\citeauthoryear{Pornpanomchai, Arpapong, Iamvisetchai, and
  Pramanus}{Pornpanomchai et~al\mbox{.}}{2011}]%
        {pornpanomchai2011thai}
\bibfield{author}{\bibinfo{person}{Chomtip Pornpanomchai},
  \bibinfo{person}{Vachiravit Arpapong}, \bibinfo{person}{Pornpetch
  Iamvisetchai}, {and} \bibinfo{person}{Nattida Pramanus}.}
  \bibinfo{year}{2011}\natexlab{}.
\newblock \showarticletitle{Thai buddhist sculpture recognition system
  (tbusrs)}.
\newblock \bibinfo{journal}{\emph{International Journal of Engineering and
  Technology}} \bibinfo{volume}{3}, \bibinfo{number}{4} (\bibinfo{year}{2011}),
  \bibinfo{pages}{342}.
\newblock


\bibitem[\protect\citeauthoryear{Saleh and Elgammal}{Saleh and
  Elgammal}{2015}]%
        {Saleh2015LargescaleCO}
\bibfield{author}{\bibinfo{person}{Babak Saleh} {and} \bibinfo{person}{Ahmed~M.
  Elgammal}.} \bibinfo{year}{2015}\natexlab{}.
\newblock \showarticletitle{Large-scale Classification of Fine-Art Paintings:
  Learning The Right Metric on The Right Feature}.
\newblock \bibinfo{journal}{\emph{CoRR}} (\bibinfo{year}{2015}).
\newblock


\bibitem[\protect\citeauthoryear{Seguin, Striolo, Kaplan, et~al\mbox{.}}{Seguin
  et~al\mbox{.}}{2016}]%
        {seguin2016visual}
\bibfield{author}{\bibinfo{person}{Benoit Seguin}, \bibinfo{person}{Carlotta
  Striolo}, \bibinfo{person}{Frederic Kaplan}, {et~al\mbox{.}}}
  \bibinfo{year}{2016}\natexlab{}.
\newblock \showarticletitle{Visual link retrieval in a database of paintings}.
  In \bibinfo{booktitle}{\emph{ECCV Workshops}}.
\newblock


\bibitem[\protect\citeauthoryear{Shamir, Macura, Orlov, Eckley, and
  Goldberg}{Shamir et~al\mbox{.}}{2010}]%
        {shamir2010impressionism}
\bibfield{author}{\bibinfo{person}{Lior Shamir}, \bibinfo{person}{Tomasz
  Macura}, \bibinfo{person}{Nikita Orlov}, \bibinfo{person}{D~Mark Eckley},
  {and} \bibinfo{person}{Ilya~G Goldberg}.} \bibinfo{year}{2010}\natexlab{}.
\newblock \showarticletitle{Impressionism, expressionism, surrealism: Automated
  recognition of painters and schools of art}.
\newblock \bibinfo{journal}{\emph{ACM Transactions on Applied Perception}}
  (\bibinfo{year}{2010}).
\newblock


\bibitem[\protect\citeauthoryear{Shimizu}{Shimizu}{2013}]%
        {faces2013}
\bibfield{author}{\bibinfo{person}{Masumi Shimizu}.}
  \bibinfo{year}{2013}\natexlab{}.
\newblock \bibinfo{booktitle}{\emph{Butsuzo no kao -Katachi to hyojo wo yomu}}.
\newblock \bibinfo{publisher}{Iwanami Shinsho}.
\newblock


\bibitem[\protect\citeauthoryear{Shozaburo}{Shozaburo}{1966}]%
        {heian1}
\bibfield{author}{\bibinfo{person}{Maruo Shozaburo}.}
  \bibinfo{year}{1966}\natexlab{}.
\newblock \bibinfo{booktitle}{\emph{Nihonchokokushikisoshiryoshusei heianjidai
  zozomeikihen}}.
\newblock \bibinfo{publisher}{Chuokoron Bijutsushuppan}.
\newblock


\bibitem[\protect\citeauthoryear{Shozaburo}{Shozaburo}{1973}]%
        {heian2}
\bibfield{author}{\bibinfo{person}{Maruo Shozaburo}.}
  \bibinfo{year}{1973}\natexlab{}.
\newblock \bibinfo{booktitle}{\emph{Nihonchokokushikisoshiryoshusei heianjidai
  juyosakuhinhen}}.
\newblock \bibinfo{publisher}{Chuokoron Bijutsushuppan}.
\newblock


\bibitem[\protect\citeauthoryear{Simonyan and Zisserman}{Simonyan and
  Zisserman}{2015}]%
        {simonyan2014very}
\bibfield{author}{\bibinfo{person}{Karen Simonyan} {and}
  \bibinfo{person}{Andrew Zisserman}.} \bibinfo{year}{2015}\natexlab{}.
\newblock \showarticletitle{Very Deep Convolutional Networks for Large-Scale
  Image Recognition}. In \bibinfo{booktitle}{\emph{3rd International Conference
  on Learning Representations, {ICLR} 2015, San Diego, CA, USA, May 7-9, 2015,
  Conference Track Proceedings}}.
\newblock


\bibitem[\protect\citeauthoryear{Soejima and Fischer}{Soejima and
  Fischer}{2008}]%
        {Guide}
\bibfield{author}{\bibinfo{person}{Hiromichi Soejima} {and}
  \bibinfo{person}{Felice Fischer}.} \bibinfo{year}{2008}\natexlab{}.
\newblock \bibinfo{booktitle}{\emph{A Guide to Japanese Buddhist Sculpture}}.
\newblock \bibinfo{publisher}{Ikeda Shoten}.
\newblock


\bibitem[\protect\citeauthoryear{Strezoski and Worring}{Strezoski and
  Worring}{2017}]%
        {strezoski2017omniart}
\bibfield{author}{\bibinfo{person}{Gjorgji Strezoski} {and}
  \bibinfo{person}{Marcel Worring}.} \bibinfo{year}{2017}\natexlab{}.
\newblock \showarticletitle{OmniArt: Multi-task Deep Learning for Artistic Data
  Analysis}.
\newblock \bibinfo{journal}{\emph{arXiv preprint arXiv:1708.00684}}
  (\bibinfo{year}{2017}).
\newblock


\bibitem[\protect\citeauthoryear{Strezoski and Worring}{Strezoski and
  Worring}{2018}]%
        {strezoski2018omniart}
\bibfield{author}{\bibinfo{person}{Gjorgji Strezoski} {and}
  \bibinfo{person}{Marcel Worring}.} \bibinfo{year}{2018}\natexlab{}.
\newblock \showarticletitle{OmniArt: A Large-scale Artistic Benchmark}.
\newblock \bibinfo{journal}{\emph{ACM Transactions on Multimedia Computing,
  Communications, and Applications (TOMM)}} \bibinfo{volume}{14},
  \bibinfo{number}{4} (\bibinfo{year}{2018}), \bibinfo{pages}{88}.
\newblock


\bibitem[\protect\citeauthoryear{Tan, Chan, Aguirre, and Tanaka}{Tan
  et~al\mbox{.}}{2016}]%
        {Tan2016CeciNP}
\bibfield{author}{\bibinfo{person}{Wei~Ren Tan}, \bibinfo{person}{Chee~Seng
  Chan}, \bibinfo{person}{Hern{\'a}n~E. Aguirre}, {and}
  \bibinfo{person}{Kiyoshi Tanaka}.} \bibinfo{year}{2016}\natexlab{}.
\newblock \showarticletitle{Ceci n'est pas une pipe: A deep convolutional
  network for fine-art paintings classification}.
\newblock \bibinfo{journal}{\emph{ICIP}} (\bibinfo{year}{2016}).
\newblock


\bibitem[\protect\citeauthoryear{Wang, He, He, Chen, and Huang}{Wang
  et~al\mbox{.}}{2019}]%
        {wang2019average}
\bibfield{author}{\bibinfo{person}{Haiyan Wang}, \bibinfo{person}{Zhongshi He},
  \bibinfo{person}{Yiman He}, \bibinfo{person}{Dingding Chen}, {and}
  \bibinfo{person}{Yongwen Huang}.} \bibinfo{year}{2019}\natexlab{}.
\newblock \showarticletitle{Average-face-based virtual inpainting for severely
  damaged statues of Dazu Rock Carvings}.
\newblock \bibinfo{journal}{\emph{Journal of Cultural Heritage}}
  \bibinfo{volume}{36} (\bibinfo{year}{2019}), \bibinfo{pages}{40--50}.
\newblock


\bibitem[\protect\citeauthoryear{Yamada}{Yamada}{2014}]%
        {reportya}
\bibfield{author}{\bibinfo{person}{Osamu Yamada}.}
  \bibinfo{year}{2014}\natexlab{}.
\newblock \showarticletitle{Chokokubunkazai ni mirareru zugakutekikaishaku}.
\newblock \bibinfo{journal}{\emph{Taikaigakujutsukoenrombunshu}}
  (\bibinfo{year}{2014}), \bibinfo{pages}{23--28}.
\newblock
\showISSN{2189-0072}


\bibitem[\protect\citeauthoryear{Yamamoto}{Yamamoto}{2006}]%
        {secret2006}
\bibfield{author}{\bibinfo{person}{Tsutomu Yamamoto}.}
  \bibinfo{year}{2006}\natexlab{}.
\newblock \bibinfo{booktitle}{\emph{Butsuzo no himitsu}}.
\newblock \bibinfo{publisher}{Asahi Shuppansha}.
\newblock


\end{thebibliography}

\end{document}